\documentclass[journal,comsoc]{IEEEtran}

\usepackage[T1]{fontenc}
\usepackage[table]{xcolor}  
\usepackage{mathtools}
\usepackage{amsmath}
\usepackage{amsmath,amssymb,amsthm}
\usepackage{graphicx}
\usepackage{subfig}
\usepackage[framemethod=tikz]{mdframed}
\usepackage{soul}
\usepackage{multirow}
\usepackage{commath}
\usepackage{url}
\usepackage{caption}
\usepackage{esvect}
\usepackage{booktabs} 
\usepackage{gensymb}
\usepackage{tablefootnote}
\usepackage{booktabs} 
\usepackage{etoolbox} 
\usepackage[vlined,boxed,commentsnumbered, ruled, linesnumbered]{algorithm2e}

\usepackage{footnote}

\newtheorem{mydef}{Definition}
\makeatletter
\def\BState{\State\hskip-\ALG@thistlm}

\makeatother

\usepackage{setspace}
\usepackage{breakcites}
\usepackage{enumerate}
\usepackage{makecell}

\usepackage{float}
\usepackage{listings}

\usepackage{tikz}
\usetikzlibrary{trees}
\usepackage{paralist}
\usepackage{ragged2e}
\usepackage[framemethod=tikz]{mdframed}
\usetikzlibrary{positioning,arrows.meta}

\definecolor{arrowblue}{RGB}{98,145,224}

\usepackage{verbatim}
\usepackage{smartdiagram}
\usepackage{metalogo}

\newcommand*{\field}[1]{\mathbb{#1}}%

\usepackage{cleveref}

\newtheorem{theorem}{Theorem}[section]

\makeatletter
\newcommand{\removelatexerror}{\let\@latex@error\@gobble}
\makeatother

\usepackage{xhfill}
\newcommand{\ditto}[1][10pt]{\xrfill{#1}~\textquotedbl~\xrfill{#1}}

\ifCLASSOPTIONcompsoc
  \usepackage[nocompress]{cite}
\else
  \usepackage{cite}
\fi

\ifCLASSINFOpdf

\else

\fi

\begin{document}

\title{Local Differential Privacy for Deep Learning}

\author{
\thanks{Manuscript received August 31, 2019; revised xx, xx;
accepted xx, xx. Date of publication xx, xx; date of
current version xx, xx. }

Pathum Chamikara Mahawaga Arachchige,
        ~Peter Bertok,
        ~Ibrahim Khalil,
        ~Dongxi Liu,
        ~Seyit~Camtepe,
        ~and~Mohammed Atiquzzaman
\IEEEcompsocitemizethanks{\IEEEcompsocthanksitem M.A.P. Chamikara is with the department of Computer Science and Software Engineering at the School of Science, RMIT University, Australia. He is also with CSIRO Data61, Melbourne, Australia.\protect\\ 
E-mail: pathumchamikara.mahawagaarachchige@rmit.edu.au
\IEEEcompsocthanksitem P. Bertok and I. Khalil are with the department of Computer Science and Software Engineering at the School of Science, RMIT University, Melbourne, Australia.
\IEEEcompsocthanksitem D. Liu and S. Camtepe are with CSIRO Data61, Sydney, Australia.
\IEEEcompsocthanksitem M. Atiquzzaman is with the School of Computer Science at the University of Oklahoma.
}
\thanks{Copyright \copyright ~20xx IEEE. Personal use of this material is permitted. However, permission to use this material for any other purposes must be obtained from the IEEE by sending a request to pubs-permissions@ieee.org.}
}

\markboth{IEEE Internet of Things Journal,~Vol.~xx, No.~xx}%
{Chamikara \MakeLowercase{\textit{et al.}}: Local Differential Privacy for Deep Learning}

\IEEEtitleabstractindextext{%
\begin{abstract}

The internet of things (IoT) is transforming major industries including but not limited to healthcare, agriculture, finance, energy, and transportation. IoT platforms are continually improving with innovations such as the amalgamation of software-defined networks (SDN) and network function virtualization (NFV) in the edge-cloud interplay. Deep learning (DL) is becoming popular due to its remarkable accuracy when trained with a massive amount of data, such as generated by IoT. However, DL algorithms tend to leak privacy when trained on highly sensitive crowd-sourced data such as medical data.  Existing privacy-preserving DL algorithms rely on the traditional server-centric approaches requiring high processing powers. We propose a new local differentially private (LDP) algorithm named LATENT that redesigns the training process. LATENT enables a data owner to add a randomization layer before data leave the data owners' devices and reach a potentially untrusted machine learning service.  This feature is achieved by splitting the architecture of a convolutional neural network (CNN) into three layers: (1) convolutional module, (2) randomization module, and (3) fully connected module. Hence, the randomization module can operate as an NFV privacy preservation service in an SDN-controlled NFV, making LATENT more practical for IoT-driven cloud-based environments compared to existing approaches. The randomization module employs a newly proposed LDP protocol named utility enhancing randomization, which allows LATENT to maintain high utility compared to existing LDP protocols. Our experimental evaluation of LATENT on convolutional deep neural networks demonstrates excellent accuracy (e.g. 91\%- 96\%) with high model quality even under low privacy budgets (e.g. $\varepsilon=0.5$). 

\end{abstract}

}

\maketitle

\begin{mdframed}[backgroundcolor=green!50,rightline=false,leftline=false]
	\centering 
	The published article can be found at \url{https://doi.org/10.1109/JIOT.2019.2952146}
\end{mdframed}

\IEEEdisplaynontitleabstractindextext

\IEEEpeerreviewmaketitle

\ifCLASSOPTIONcompsoc
\IEEEraisesectionheading{\section{Introduction}\label{sec:introduction}}
\else
\section{Introduction}
\label{sec:introduction}
\fi

\IEEEPARstart{T}{he}  internet of things (IoT) has become one of the essential assets that transform many industries, such as oil and gas, healthcare, agriculture, finance, and transportation. The production of a large amount of IoT data has led to the advancement of different technologies, such as big data analytics and machine learning (ML), and has opened up new opportunities.  The underlying technologies of IoT are continuously improving to address complex heterogeneity and to improve programmable flexibility. Network infrastructural improvements, such as the amalgamation of software-defined networks (SDN) and network function virtualization (NFV) in the edge-cloud interplay, are promising better quality of service (QoS) for complex IoT-driven applications. Such environments can be fully utilised using ML to generate efficient and advanced analytics. However, the server-centric architectures employed by many ML algorithms limits their integration into distributed environments such as SDN-controlled NFV.

Compared to traditional machine learning approaches, deep learning  (DL) shows remarkable success in addressing complex problems such as image classification, natural language processing, and speech recognition.  DL models are often trained on sensitive crowd-sourced data such as personal images, health records, and financial records. When DL models are trained on massive databases containing sensitive data, they tend to expose private information~\cite{abadi2016deep,shokri2017membership}. With the advancement of distributed, cloud-based machine learning environments such as those offered by Google and Amazon ~\cite{abadi2016tensorflow,low2012distributed}, more users may become vulnerable to such attacks. Trusting these environments, users may feed their data to train the models and obtain white-box or black-box access to these models without being concerned about the actual training process. However, an adversary can easily implement malicious algorithms and offer them as part of the training process.  Malicious algorithms may memorize the sensitive user information as part of the trained models. Adversaries can later extract and approximate the memorized information, and thereby obtain information about the users and breach their privacy~\cite{song2017machine}. Privacy inference attacks, such as membership inference, show the vulnerability of deep learning models trained on sensitive data even when they are released as black-box models~\cite{shokri2017membership}. Another example that shows the weakness of trained ML models is model inversion attacks that recover images from a facial recognition system~\cite{fredrikson2015model}. It is essential that machine learning as a service employs sufficient privacy-preserving mechanisms to limit privacy leaks of trained DL models.  It is also essential that these privacy-preserving approaches for DL can be used for IoT based applications such as smart healthcare, IIoT, and Industry 4.0.

\begin{figure}[H]
	\centering
	\scalebox{0.52}{
	\includegraphics[width=0.6\textwidth, trim=0.5cm 0cm 0.5cm
	 0cm]{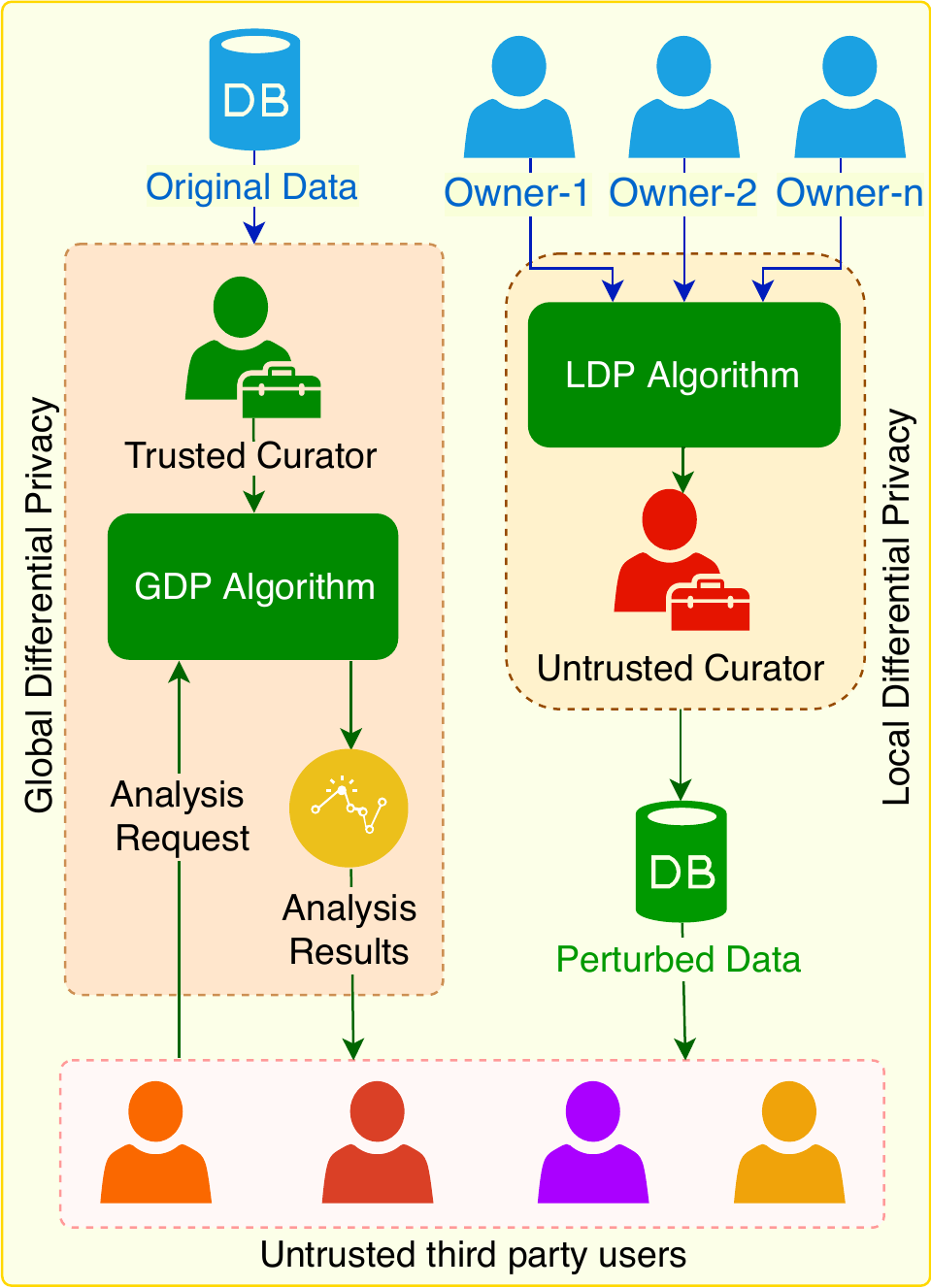}
	 }
	\caption{Global vs. Local differential privacy}
	\label{glvsdp}
\end{figure}

In this paper, we examine the privacy issues of deep learning and develop a distributed privacy-preserving mechanism using differential privacy (DP) to control and limit privacy leaks in deep learning.  DP constitutes a robust framework guaranteeing strong levels of privacy ~\cite{dwork2014algorithmic}. The existing benchmark privacy-preserving approaches for deep learning are based on global differential privacy (GDP)~\cite{shokri2015privacy,abadi2016deep}. However, we chose local differential privacy (LDP) over GDP. As shown in Fig. \ref{glvsdp}, GDP employs a trusted curator to apply calibrated noise to produce differential privacy ~\cite{xiao2008output, kairouz2014extremal}. The necessity of a trusted curator makes the existing GDP methods unsuitable for practical DL services, such as those offered by Google. In such a scenario, the GDP algorithm should reside in the server, and the original data need to be uploaded to the server for training. This approach can pose a threat to privacy, as an adversary can perform server-centric attacks such as membership inference and model memorizing attacks. Moreover, DL algorithms are inherently computationally complex, and privacy-preserving solutions on DL models also tend to be complex and need high computational processing power. As a result, GDP algorithms are preferred to run on high-performance computers, and resource-constrained data owners can not use them in untrusted environments. Furthermore, noise calibration of GDP methods, such as Laplacian and Gaussian mechanisms for DL models, can be complex, indefinite, and produce less accurate results or entail a higher level of privacy leak~\cite{shokri2015privacy,abadi2016deep}.   However, in LDP, data owners perturb their data before releasing them, which avoids the need for a trusted third party while guaranteeing better privacy, as depicted in Fig. \ref{glvsdp}~\cite{kairouz2014extremal}.  The local approach to data perturbation in LDP innovates the development of distributed privacy preservation algorithms for many modern distributed scenarios, such as those based on IoT.

Our contribution is a distributed LDP mechanism with a new LDP protocol for limiting the privacy leaks of convolutional neural network (CNN) models that are released as black-box models. The proposed algorithm (named LATENT) employs the properties of randomized response~\cite{fox2015randomized}, a popular survey technique that satisfies local differential privacy. The LDP setting and the layered architecture of LATENT allow privacy-preserving communication between several parties, which is not possible with existing GDP methods for deep learning. We conducted an in-depth analysis of the existing LDP protocols and devised a new protocol named utility enhancing randomization (UER) that provides better utility than existing LDP protocols.  We first improved the optimized unary encoding protocol (OUE) to propose a new LDP protocol named modified OUE  (MOUE), which provides enhanced flexibility of binary string randomization. OUE is an LDP protocol that follows the intuition of randomizing 1's and 0's differently to improve utility. MOUE achieves improved flexibility by introducing an additional coefficient $\alpha$ (named as the privacy budget coefficient) that provides improved flexibility in choosing randomization probabilities. We then followed the motivation behind MOUE to propose UER that maintains the utility during the randomization of long binary strings with high sensitivity. LATENT can be easily integrated into modern environments such as SDN controlled NFV by moving the layer of randomization to run as an NFV service. As the LDP approach of LATENT enables the control of the privacy budget before the perturbation process, accuracy can be effectively tuned independently. In other words, LATENT reduces the impact of the privacy budget ($\varepsilon$) on the accuracy, and this leads to significantly higher privacy and accuracy than what is offered by existing solutions.  Compared to current GDP methods for deep learning, LATENT provides excellent accuracy (above 90\%) under extreme cases of privacy budgets (e.g. $\varepsilon=0.5$) that ensure minimum leak. Our experiments clearly show that a general-purpose computer is sufficient to perform the required computations efficiently and reliably at the data owner's end. Accordingly, LATENT can be a more practical and robust tool to limit the privacy leak of deep learning models than existing methods.


The rest of the paper is organized as follows. The underlying concepts used in LATENT are presented in Section \ref{background}.  Section \ref{proplatent} explains the steps of the differentially private mechanism for deep learning. The results of LATENT  are discussed in Section \ref{resdis}. Section \ref{relwork} provides a summary of existing related work.  The paper is concluded in Section \ref{conclusion}.

\section{Background}
\label{background}
This section provides brief descriptions of the underlying concepts of LATENT.  It includes brief summaries of basic principles related to "Differential Privacy"  and to "Deep Learning", which are used in LATENT.

\subsection{Differential Privacy}

Differential privacy (DP) is a privacy model that is known to render maximum privacy by minimizing the chance of individual record identification~\cite{chamikara2019efficient}. In principle, DP defines the bounds to how much information can be revealed to a third party/adversary about someone's data being present in a particular database. Conventionally $\varepsilon$ (epsilon) and $\delta$ (delta) are used to denote these bounds, which describe the level of privacy rendered by a randomized privacy preserving algorithm ($M$) over a particular database ($D$).

\subsubsection{Privacy budget / privacy loss ($\varepsilon$)}
$\varepsilon$ is called the privacy budget that provides an insight into the privacy loss of a DP algorithm. The higher the value of $\varepsilon$, the higher the privacy loss. 

\subsubsection{Probability to fail / probability of error ($\delta$)}
$\delta$ is the parameter that accounts for "bad events" that might result in high privacy loss; $\delta$ is the probability of the output revealing the identity of a particular individual, which can happen $\delta\times n$ times where $n$ is the number of records. 
To minimize the risk of privacy loss, $\delta \times n$ has to be maintained at a low value.  For example, the probability of a bad event is 1\% when $\delta=1/{100\times n}$.

\subsubsection{Definition of differential privacy}

Let's take a dataset $D$ and two of its adjacent datasets, $x$ and $y$, where $y$ differs from $x$ only by one person. Assume,  datasets $x$ and $y$ as being collections of records from a universe $\mathcal{X}$ and $\mathbb{N}$ denotes the set of all non-negative integers including zero.  Then $M$ satisfies ($\varepsilon$, $\delta$)-differential privacy if Equation \eqref{dpeq} holds.

\begin{mydef}
A randomized algorithm $M$ with domain $\mathbb{N}^{|\mathcal{X}|}$  and
range $R$: is ($\varepsilon$, $\delta$)-differentially private for  $\delta \geq 0$  if for every adjacent datasets $x$, $y$ $\in$ $\mathbb{N}^{|\mathcal{X}|}$
 and for any subset $S \subseteq R$, 
\end{mydef}

\begin{equation}
Pr[(M(x) \in S)] \leq \exp(\varepsilon) Pr[(M(y) \in S)] + \delta
\label{dpeq}
\end{equation}

\subsection{Global vs. Local Differential Privacy}

Global differential privacy (GDP) and local differential privacy (LDP) are two approaches that can be used by randomized algorithms to achieve differential privacy. As depicted in Fig. \ref{glvsdp}, GDP employs a trusted curator who applies carefully calibrated random noise to the real values returned for a particular query. 
The most frequently used noise generation processes for GDP  include Laplace mechanism and Gaussian mechanism~\cite{dwork2014algorithmic}.  The GDP setting is also called the trusted curator model~\cite{chan2012differentially}. A randomized algorithm, $M$ provides $(\varepsilon, \delta)$-global differential privacy if Equation \eqref{dpeq} holds.  
LDP needs no trusted third party, hence it is also called the untrusted curator model ~\cite{kairouz2014extremal}. With LDP, data is randomized before the curator can access it. 
LDP can also be used by a trusted party to randomize all records in a database at once. The right-hand column of Fig. \ref{glvsdp} represents the LDP setting.   LDP algorithms may often produce too noisy data, as noise is applied commonly to achieve individual data privacy. LDP is considered to be a strong and rigorous notion of privacy that provides plausible deniability. Due to the above properties, LDP is deemed to be a state-of-the-art approach for privacy-preserving data collection and distribution. A randomized algorithm $A$ provides $\varepsilon$-local differential privacy if Equation \eqref{ldpeq} holds ~\cite{erlingsson2014rappor}.

\begin{mydef}
A randomized algorithm $A$  satisfies $\varepsilon$-local differential privacy if for all pairs of client's values $v_1$ and $v_2$ and for all $Q \subseteq Range(A)$ and for ($\varepsilon \geq 0$), Equation \eqref{ldpeq} holds. $Range(A)$ is the set of all possible outputs of the randomized algorithm $A$.
\end{mydef}

\begin{equation}
Pr[A(v_1) \in Q] \leq \exp(\varepsilon) Pr[A(v_2) \in Q]
\label{ldpeq}
\end{equation}

\subsection{Randomized Response}

Randomized response is a survey technique to eliminate evasive answer bias by randomizing the responses to a survey question with the answer "yes" or "no"~\cite{warner1965randomized}. An answer is randomized by flipping two independent, unbiased coins.  The answer is truthful if the first coin comes up "heads", else, the second coin is flipped, and the answer is "yes"  if "heads", "no" if "tails". Assume that a biased coin is used and the probability of a user providing an answer truthfully is $p$ (otherwise provides the opposite of the true answer, with $(1-p)$ probability). It has been shown that  this approach provides $\varepsilon$- differential privacy when $p=e^{\varepsilon}/(1+e^{\varepsilon})$~\cite{kairouz2014extremal}.

\subsection{Sensitivity, Privacy Budget ($\varepsilon$), and Determination of the Probability ($p$) of Randomization}
\label{rappor}
To quantify the probability of randomization ($p$, the probability of preserving an original bit) of an LDP process that is based on transferring bit strings, we can use the method employed by Randomized Aggregatable Privacy-Preserving
Ordinal Response (RAPPOR), which is an LDP algorithm proposed by Google~\cite{erlingsson2014rappor}. RAPPOR is motivated by the problem of estimating a client-side distribution of string values drawn from a discrete data dictionary. One application of RAPPOR is to track the distribution of users' browser configuration strings in the Chrome web browser. 

Sensitivity is defined as the maximum influence that a single individual can have on the result of a numeric query. Consider an arbitrary function $f$, the sensitivity  $\Delta f$ of $f$ can be given as in Equation \eqref{seneq} where x and y are two neighboring datasets and $\lVert . \rVert_1$ represents the $L1$ norm of a vector~\cite{wang2016using}. 

\begin{equation}
\Delta f=max\{\lVert f(x)-f(y) \rVert_1\}
\label{seneq}
\end{equation}
Since RAPPOR is an LDP algorithm, it considers $x$ and $y$ to be a pair of adjacent inputs in RAPPOR's definition of global sensitivity. In RAPPOR any input $v_i$ is encoded as a vector of $d$ bits, and each $d$-bit vector contains $d-1$ zeros and $1$ one, so the maximum difference, $\Delta f$ (the sensitivity) between two adjacent vectors, is 2 bits. In other words, the underlying data representation of RAPPOR, $f$ has a sensitivity of 2. RAPPOR is an LDP algorithm when the probability $p$ of preserving the true value of an original bit in randomization follows Equation \eqref{rapeq}, where $\varepsilon$ is the privacy budget offered by the LDP process~\cite{erlingsson2014rappor,qin2016heavy}. 

\begin{equation}
p=\frac{e^{\frac{\varepsilon}{\Delta f}}}{1+e^\frac{\varepsilon}{\Delta f}}=\frac{e^{\frac{\varepsilon}{2}}}{1+e^\frac{\varepsilon}{2}}
\label{rapeq}
\end{equation}

\subsection{Properties of Differential Privacy}
Postprocessing invariance/robustness, quantifiability, and composition are three of the essential characteristics of differential privacy~\cite{bun2016concentrated}. Although additional computations are carried out on the outcome of a differentially private algorithm, they do not weaken the privacy guarantee. So, the results of additional computations on $\varepsilon$-DP outcome will still be $\varepsilon$-DP. This property of DP is called the postprocessing invariance/robustness. Quantifiability is the ability of DP scenarios to provide transparency in calculating the precise amount of perturbation applied by a particular randomization process. Thus, the user of a particular DP algorithm knows the level of privacy provided by the data/results released after perturbation.   Composition is the degradation of privacy when multiple differentially private algorithms are performed on the same or overlapping datasets~\cite{bun2016concentrated}. According to DP definitions, when two DP algorithms $\varepsilon_1$-DP and $\varepsilon_2$-DP are applied to the same or overlapping datasets, the union of the results is equal to $(\varepsilon_1+\varepsilon_2)$-DP~\cite{bun2016concentrated}. The more DP algorithms are applied to the same data, the more privacy loss is accumulated. Depending on the process of synthesis, DP algorithms can be categorized into the two types; basic algorithms or derived algorithms~\cite{chanyaswad2018mvg}.  Differential privacy is self-contained in basic algorithms while the derived algorithms are derived from existing methods by applying the theories of composition and postprocessing invariance.

\begin{figure}[H]
	\centering
	\scalebox{0.47}{
	\includegraphics[width=1\textwidth, trim=0.5cm 0cm 0.5cm
	 0cm]{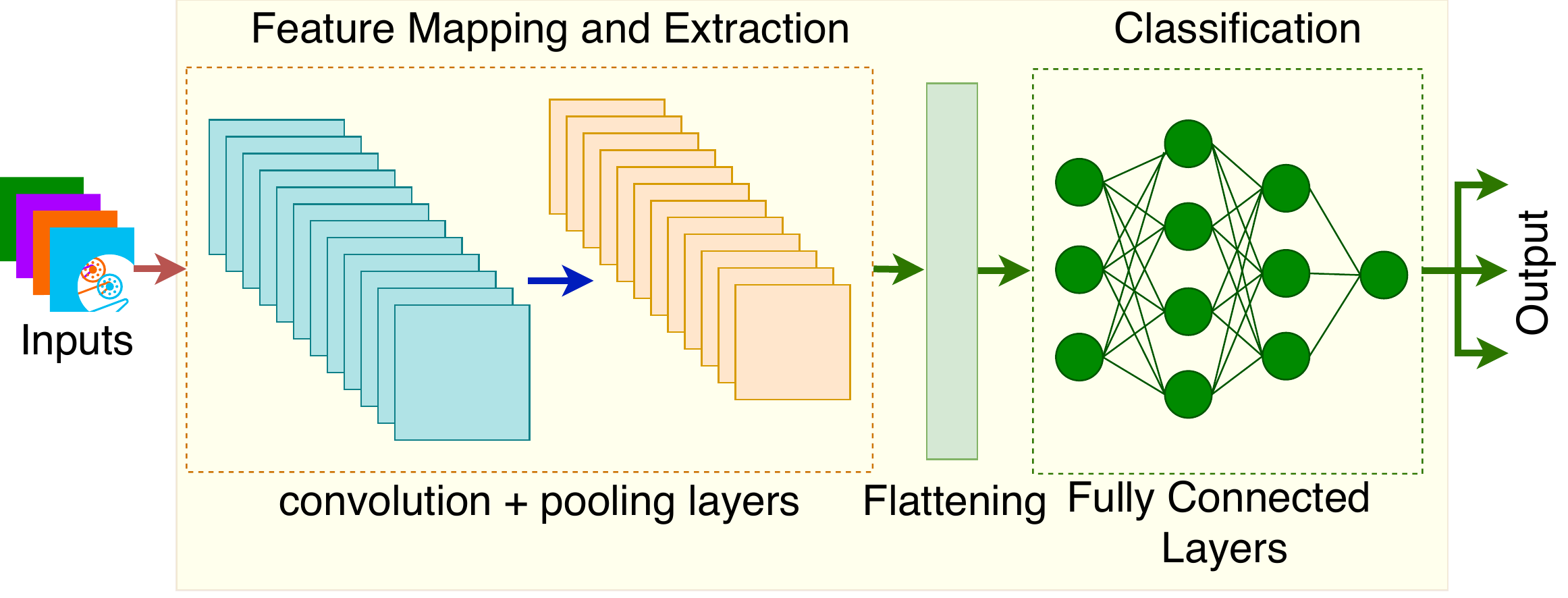}
	 }
	\caption{Generic architecture of a convolutional neural network}
	\label{convarchi}
\end{figure}

\subsection{Deep Learning Using Convolutional Neural Networks}
\label{deeplearning}

A CNN is commonly trained to recognize essential features of images. As shown in Fig. \ref{convarchi}, a CNN uses a collection of layers named convolution layers with large receptive fields.  A sequence of steps through this stack of convolution layers is followed by an intermediate functionality called pooling to reduce the dimensions from the previous layer to the next layer. The final pooled output which is produced from the last convolution layer is flattened to produce a sizeable 1-D vector~\cite{schmidhuber2015deep}. Then a fully connected artificial neural network (ANN) is trained using these input vectors to generate predictions (classifications) on the inputs (images). An ANN is more or less a connected network of processing modules called neurons, each producing a sequence of real-valued activations.

Overfitting happens when the training accuracy is significantly higher than the testing accuracy~\cite{schmidhuber2015deep, kim2014convolutional}. Good quality models avoid overfitting. Regularization, image augmentation, and hyperparameter tuning are three of the commonly used concepts to avoid overfitting and improve the performance and robustness of neural networks~\cite{schmidhuber2015deep, kim2014convolutional}. Regularization is the process of applying any modification to a learning algorithm to reduce the generalization error. Regularization can be achieved using dropouts, where a certain percentage of neurons are randomly dropped in each epoch (training cycle) to avoid overfitting. Image augmentation is a data preparation technique, which uses the existing input images in the training dataset and manipulates them to create many altered versions of the same input using different transformation methods such as reflection, sheer, and rotation. This technique allows an ANN to learn a wider variety of inputs to make the trained model more generalizable with high robustness~\cite{krizhevsky2012imagenet}.  In hyperparameter tuning the inputs to hyperparameters such as percentage dropouts, batch size, activation functions, number of neurons, number of epochs, and optimizer are changed under different training phases to identify the best case study that returns the best results~\cite{schmidhuber2015deep, kim2014convolutional}. Batch size is the number of training examples that are going to be propagated in one forward/backward pass~\cite{kim2014convolutional}. Activation functions define the output of a particular neuron given a set of inputs that, with corresponding weights, introduce non-linear properties to the network~\cite{schmidhuber2015deep}. A neuron (also called a node) is the primary component of an artificial neural network. A single pass in which the entire dataset is introduced forward and backward through the neural network is called an epoch~\cite{schmidhuber2015deep}. An optimizer (or an optimization algorithm) is used to update the model parameters such as weights and bias values~\cite{schmidhuber2015deep}

\subsection{Amalgamation of SDN and NFV in Edge-Cloud Interplay}
SDN and NFV are two types of programmable infrastructures that improve the versatility of networking. Both technologies work based on the concept of creating virtual instances for network functionalities where SDN virtualizes controlling aspects, and NFV virtualizes essential network functions such as encrypting channels. The amalgamation between SDN and NFV can bring forward many advanced capabilities in terms of flexibility, efficiency, and quality of service with increased reconfigurability. SDN-controlled NFV can introduce a series of virtualizations that can benefit the edge-cloud interplay, which can improve the security and quality of service of communication between local devices and cloud servers~\cite{vaquero2014finding,kaur2019energy}. 

\section{Our Approach: LATENT}
\label{proplatent}
This section discusses the differentially private mechanism employed in LATENT for deep learning. LATENT can be classified as a derived differentially private algorithm which is based on the randomized response technique. LATENT uses two properties of differential privacy: postprocessing invariance and composition when applying differential privacy to a CNN. LATENT uses regularization, image augmentation, and hyperparameter tuning to optimize its performance under noisy input conditions in the randomization process. We implemented and tested LATENT on convolutional neural networks using the Python Keras neural networks API, which runs on top of the TensorFlow dataflow engine developed by Google~\cite{abadi2016tensorflow,chollet2015keras}.  Keras provides a high-level neural-network API designed primarily for fast experimentation.

\subsection{Introduction of the Intermediate Layer (LATENT) to Inject Differential Privacy into the CNN Architecture}
\label{convolearn}

\begin{figure}[ht]
    \centering
    \scalebox{1}{
    \includegraphics[width=0.47\textwidth, trim=0.5cm 0cm 0.5cm
     0cm]{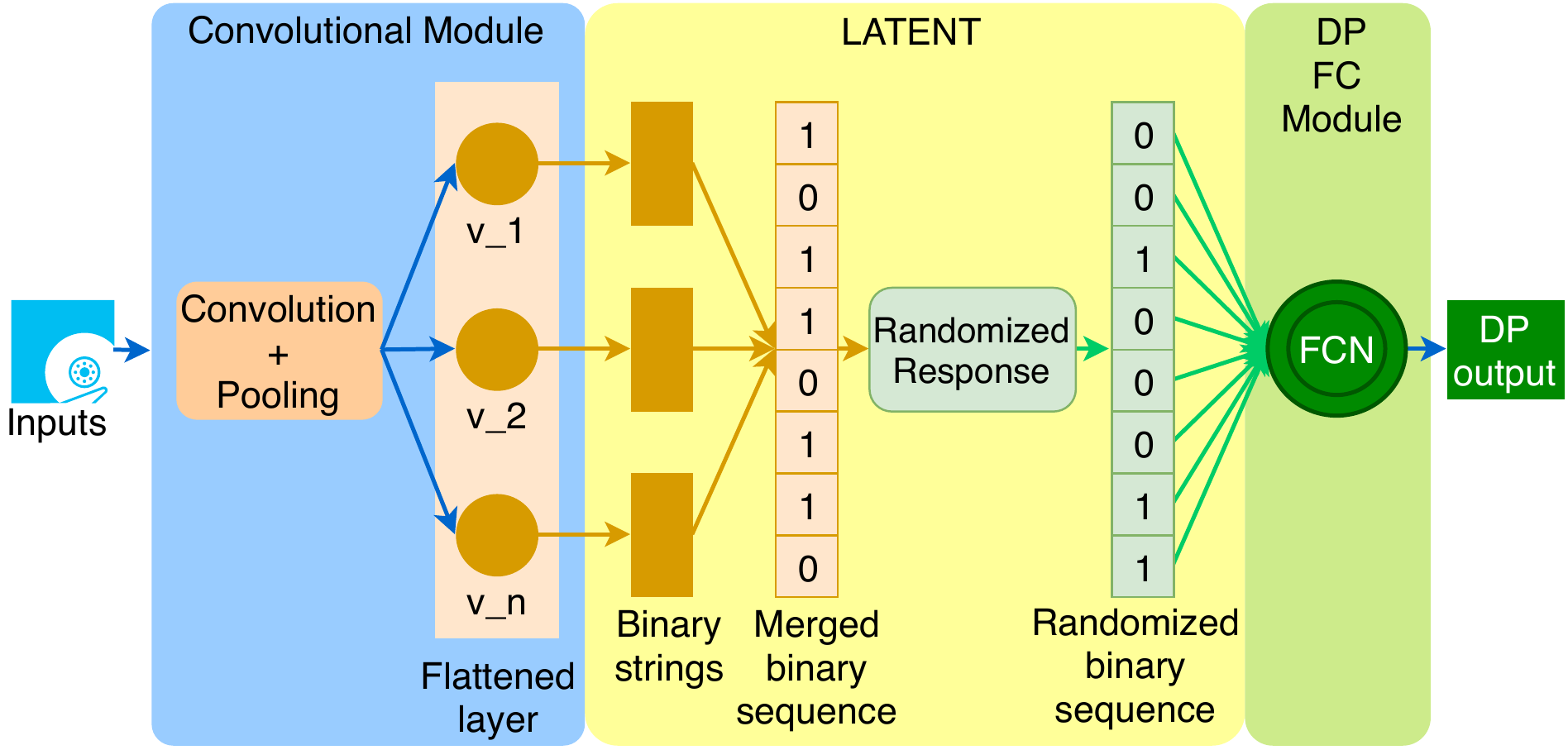}
     }
    \caption{CNN architecture with the LATENT randomization layer (FCN: fully connected network)}
    \label{latentcnn}
\end{figure}

As shown in Fig. \ref{latentcnn}, we divide the structure of a convolutional neural network into two main modules, and introduce an intermediate module of randomization into the CNN structure. Recall (described in Section \ref{deeplearning}) that in CNN, the input features are initially subjected to dimensionality reduction, using a collection of convolutional layers and pooling layers. The output of the final pooling layers is flattened into a single 1-D array before feeding it to a fully connected artificial neural network. We call this part of a CNN the convolutional module, and we name the fully connected network component of the CNN as the FC module. We insert the randomization layer named LATENT between the convolutional module and the FC module, as shown in Fig. \ref{latentcnn}. In the proposed architecture we use the convolutional module only to generate the 1-D flattened output that corresponds to a particular image input. This flattened output is simply a one-dimensional column vector of float values (real-valued numbers).

\subsubsection{Apply z-score normalization to LATENT's input values}

\label{latentzsense}
LATENT converts the input values to binary values before randomization. The inputs can have different ranges. Conversion of large values or small fractions into binary can involve a large number of bits. This can introduce an inconsistent level of complexity to the algorithm. To avoid this complexity, we apply z-score normalization to the values of the 1-D vector coming from the flattening layer.

\subsubsection{Define the bounds (lengths of the segments) for the binary conversion}
The length of the bit pattern establishes the range of a particular z-score normalized input. The upper bound and the lower bound of a specific input need to be initially estimated. Fig. \ref{ftobinary} shows the arrangement of bits of the binary conversion of a z-score normalized input. As shown in the figure, there are three primary segments of the binary string. The first bit represents the sign of the input (1 for negative and 0 for positive). The other two parts are for the whole number and the fraction part of an input number respectively. Selection of the number of bits for the whole number depends on the maximum value of the whole number that needs to be represented.  Due to z-score normalization, the number of bits necessary to represent the whole number is small.  Selection of the number of bits for the fraction depends on the precision (how close is the binary fraction's decimal value to the input's fraction value).  For more precision, a higher number of bits needs to be used for the fraction.

\subsubsection{Convert each value of the flattened layer to binary using the bounds}
\label{binconversion}
\begin{figure}[H]
	\centering
	\scalebox{1}{
	\includegraphics[width=0.40\textwidth, trim=0.5cm 0cm 0.5cm
	 0cm]{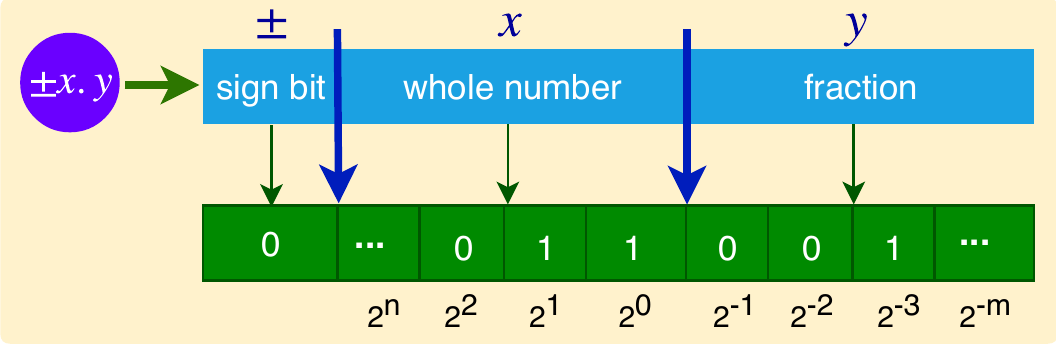}
	 }
	\caption{Direct mapping of a float/integer to binary}
	\label{ftobinary}
\end{figure}

After determining the length of the components of binary strings of the inputs, the inputs can be mapped as shown in Fig. \ref{ftobinary}. The figure shows the direct mapping of an integer/float value to its binary representation. The binary representation can be generated according to Equation \eqref{intbinary}, where $n$ and $m$ are the numbers of binary digits of the whole number and the fraction respectively, $x$ represents the original input value where $x\in \mathbb{R}$, and  $g(i)$ represents the $i^{th}$ bit of the binary string where the least significant bit is represented when $k=-m$. The sign bit is 1 for negative values and 0 for positive values. The sign bit is assigned to the most significant bit of the binary string. 

\begin{equation}
g(i)=\,{\Big(\left\lfloor 2^{-k}\, \abs x \right\rfloor\text{ mod }2\Big)_{k=-m}^{n}}\ \text{where, } i=k+m
\label{intbinary}
\end{equation}

\subsubsection{Merge the binary strings to reduce the privacy loss}

We merge all the binary strings into one long binary string to avoid privacy loss due to the composition property of differential privacy. If we conducted the randomization on each binary string corresponding to the flattened 1-D vector separately, it would add up the privacy budgets of all the randomization steps. If $r$ binary strings were randomized, the resulting privacy loss of the final randomization would be $r\times \varepsilon$.  As LATENT conducts the randomization on a particular merged binary string at once, we can maintain the privacy loss at the input value of $\varepsilon$.

\subsubsection{Define the probability of randomization ($p$) in terms of $\varepsilon$}
\label{probsect}

The probability of randomization $p$ (i.e. the probability of preserving the true value of an original bit) is calculated in terms of the privacy budget ($\varepsilon$) before the randomization of the merged binary string. Recall that in the randomized response technique (as described in Section \ref{rappor}) used in RAPPOR,  when the difference of the number of bits of two neighboring inputs is $d$, the sensitivity becomes $d$. In the case of LATENT, the length of a binary string is $l=(n+m+1)$ which makes the length of the merged binary string equal to $l\times r$ where $r$ is the number of outputs of the flattening layer of the convolutional module.  According to our method of binary conversion, two consecutive inputs can differ by at most of $l\times r$ bits.  Consequently, LATENT has a sensitivity of $l\times r$. Now,  we can represent the probability of randomization according to Equation \eqref{mergedrandprob}. We note that merging the binary strings together increases sensitivity, hence, makes increasing the amount of randomization necessary.

\begin{equation}
p=\frac{e^{\varepsilon/rl}}{1+e^{\varepsilon/rl}}
\label{mergedrandprob}
\end{equation}

\subsubsection{Modifying optimized unary encoding to improve utility}
As discussed in Section \ref{probsect}, the probability of randomization in reporting opposite of the true bits is $(1-p) = \frac{1}{1+e^{\varepsilon/rl}}$. However, this can introduce an unreliable level of randomization to the output of the LATENT layer, as the sensitivity = $rl$ is extensive, as discussed in Section \ref{probsect}. To improve the utility, we follow the intuition behind  Optimized Unary Encoding (OUE)~\cite{wang2017locally}. OUE perturbs 0 and 1 differently to reduce the probability of perturbing 0 to 1 ($p_{0\rightarrow 1}$) as there are considerably more 0's than 1's when the input binary string is long. We propose a new approach to further optimize the selection of probabilities of randomization, which can provide enhanced utility when the bit strings are long.

Let $v_i$ represent an instance in the database and $B$ is a $d$ bit binary encoded version of $v_i$. $B[i]$ represents the $i^{th}$ bit and $B^{\prime}[i]$ is the perturbed $i^{th}$ bit.  Assume that, only the $j^{th}$ position of $B$ is set to 1, whereas the other bits are set to zero. Unary Encoding (UE)~\cite{erlingsson2014rappor} perturbs the bits of $B$ according to Eq. \ref{pereq11}. 

\begin{equation}
\operatorname{Pr}\left[B^{\prime}[i]=1\right]=\left\{\begin{array}{ll}{p,} & {\text { if } B[i]=1} \\ {q,} & {\text { if } B[i]=0}\end{array}\right.
\label{pereq11}
\end{equation}

UE satisfies $\varepsilon$-LDP~\cite{erlingsson2014rappor,wang2017locally} for,

\begin{equation}
\varepsilon=\ln \left(\frac{p(1-q)}{(1-p) q}\right)
\label{pqratio}
\end{equation}

This can be proven (refer to Appendix \ref{appdx1}) as done in ~\cite{erlingsson2014rappor,wang2017locally} for any inputs $v_1$, $v_2$, and $B$ with sensitivity = 2. Utilizing the concept of UE, optimized unary encoding (OUE) in its optimal setting sets $p=\frac{1}{2}$ and $q=\frac{1}{1+e^\varepsilon}$, so that the randomization improves the budget allocation for transmitting the 0 bits in their original state as much as possible. According to Eq.\ref{pqratio}, we can show that OUE provides $\varepsilon$-LDP when $p=\frac{1}{2}$, $q=\frac{1}{1+e^\varepsilon}$, and  sensitivity = 2 (refer to Appendix \ref{appdxoue} for proof).

When the sensitivity is high, the bit strings will have many 1s. In such a scenario (such as in LATENT), more flexibility for controlling the randomization of 1s is also essential to generate better utility. Following this intuition, we propose the Upper Bound (UB) Theorem \ref{ubtheorem}. In UB theorem, we consider a coefficient ($\alpha$, the privacy budget coefficient) during the probability selection. This model provides $\varepsilon_{ub}$-LDP where $\varepsilon_{ub}=ln{(\alpha^2 e^{\varepsilon})}$.

\begin{theorem}
\label{ubtheorem}
\textbf{(Upper bound (UB) theorem)}\\
Let $UB(\varepsilon)$ be the upper bound of the privacy budget when the sensitivity = 2. Let the perturbation probabilities, $p = \frac{\alpha e^{\frac{\varepsilon}{2}}}{1+\alpha e^{\frac{\varepsilon}{2}}}$ and $q = \frac{1}{1+\alpha e^{\frac{\varepsilon}{2}}}$. Then $UB (\varepsilon)$) = $\ln{(\alpha^2 e^{\varepsilon})}$, where $\alpha$ is the privacy budget coefficient (refer to Appendix \ref{upbproof} for proof). 
\end{theorem}

We can extend the idea in the UB theorem and use the privacy budget coefficient ($\alpha$) to modify OUE as defined in Theorem \ref{theoremmoue}; $\alpha$ provides more flexibility to MOUE in choosing the randomization probabilities. By increasing $\alpha$, we can increase the probability of transmitting the 0 bits in their original state. 

\begin{theorem}
\label{theoremmoue}

\textbf{(Modified OUE (MOUE))}\\
For any inputs $v_1$,$v_2$ in  MOUE, $\operatorname{Pr}\left[\boldsymbol{B}\left[v_{1}\right]=1 | v_{1}\right] = \frac{1}{1+\alpha}$, $\operatorname{Pr}\left[\boldsymbol{B}\left[v_{1}\right]=1 | v_{2}\right]= \frac{\alpha}{1+\alpha}$, $\operatorname{Pr}\left[\boldsymbol{B}\left[v_{2}\right]=0 | v_{1}\right] = \frac{\alpha e^{\varepsilon}}{1+\alpha e^{\varepsilon}} $, $\operatorname{Pr}\left[\boldsymbol{B}\left[v_{2}\right]=0 | v_{2}\right]= \frac{1}{1+\alpha e^{\varepsilon}}$. Then, MOUE provides $\varepsilon$-LDP (refer to Appendix \ref{proofmoue} for proof).
\end{theorem}

MOUE is suitable for randomizing bit strings, which has a considerable number of 1s along with a large number of 0s such as in LATENT. Theorem \ref{mouelarge} defines the probabilities for cases where the sensitivity is considerably large. Preserving as much 0s as possible will allow the utility to be preserved while tolerating an anticipated loss in the probability of preserving the original 1.

\begin{theorem}
\label{mouelarge}
\textbf{(MOUE for high sensitivities)}\\
MOUE for LATENT provides $\varepsilon$-LDP, when $\operatorname{Pr}\left[\boldsymbol{B}\left[v_{1}\right]=1 | v_{1}\right] = \frac{1}{1+\alpha}$, $\operatorname{Pr}\left[\boldsymbol{B}\left[v_{1}\right]=1 | v_{2}\right]= \frac{\alpha}{1+\alpha}$, $\operatorname{Pr}\left[\boldsymbol{B}\left[v_{1}\right]=1 | v_{2}\right] = \frac{\alpha e^{\frac{\varepsilon}{rl}}}{1+\alpha e^{\frac{\varepsilon}{rl}}} $, $\operatorname{Pr}\left[\boldsymbol{B}\left[v_{2}\right]=0 | v_{2}\right]= \frac{1}{1+\alpha e^{\frac{\varepsilon}{rl}}}$ (refer to Appendix \ref{mohsen} for proof). 
\end{theorem}

\subsubsection{Improving the utility of randomized binary strings}
Although MOUE improves the utility of randomized output, it can also massively increase the randomization of 1's when $\alpha$  is large. MOUE improves the probabilities of randomization by randomizing 1 and 0 differently following the intuition behind OUE. By further extending this idea, we apply two randomization models over the bits of a binary string to enhance the utility of randomized binary strings. In this way, we try to randomize half of the bits in the bit string differently compared to the other half as defined in Theorem \ref{uertheorem}.  When $\alpha$ is increased, UER will increase the probability of preserving 0s in their original state. However, the probability of preserving 1s in their original state increases for half of the string while the corresponding probability decreases for the other half of the string. In this way, UER maintains the privacy budget at $\epsilon$.

\begin{theorem}
\label{uertheorem}
\textbf{(Utility enhancing randomization (UER))}\\
Let $p(B[i]v)$ be the probability of randomizing the $i^{th}$ bit of the binary  encoded string of $v$. For any inputs $v_1, v_2$ with a sensitivity = $rl$, define the probability, $p(B[i]v)$ as in Eq. \ref{twomodprob}. Then the randomization provides $\epsilon$-LDP (refer to Appendix \ref{proofuer} for proof).

\begin{equation}
p(B[i]v)=\left\{\begin{array}{ll}{\operatorname{Pr}\left[\boldsymbol{B}\left[v_{1}\right]=1 | v_{1}\right] = \frac{\alpha}{1+\alpha}} & {\text { if } i \in 2n; n\in \field{N}} \\
{\operatorname{Pr}\left[\boldsymbol{B}\left[v_{2}\right]=0 | v_{1}\right] = \frac{\alpha e^{\frac{\varepsilon}{rl}}}{1+\alpha e^{\frac{\varepsilon}{rl}}}} & {\text ~~~~~~~ \ditto} \\ {\operatorname{Pr}\left[\boldsymbol{B}\left[v_{1}\right]=1 | v_{1}\right] = \frac{1}{1+\alpha^3}} & {\text { if } i \in 2n+1} \\ {\operatorname{Pr}\left[\boldsymbol{B}\left[v_{2}\right]=0 | v_{1}\right] = \frac{\alpha e^{\frac{\varepsilon}{rl}}}{1+\alpha e^{\frac{\varepsilon}{rl}}}} & {\text ~~~~~~~ \ditto }\end{array}\right.
\label{twomodprob}
\end{equation}
\end{theorem}

\subsubsection{Conduct UER on the bits of the merged binary strings}
\label{randomization}

Each bit in the merged binary string is subjected to  MOUE, given in Equation \eqref{mergedrandprob}. The higher the $p$ is, the lower the randomization of the binary string will be. According to Equation \eqref{mergedrandprob}, higher $\varepsilon$ values and lower $l\times r$ values will result in higher $p$ values. In LATENT, $l\times r$ is a considerably larger value than $\varepsilon$, $p$ is often a smaller value.  MOUE allows LATENT to fine-tune the probabilities of randomization by calibrating $\alpha$. By increasing $\alpha$ to be greater than 5, MOUE decreases the probability of perturbing 0 to 1. This feature of MOUE  helps LATENT to maintain the utility as there are a large number of 0's compared to 1's in binary encoded inputs of LATENT. However, this reduces the probability of releasing the real bits when the input bit is 1 for half of the bits in the input bit strings. Compared to RAPPOR, a binary string in LATENT has many 1's. As a result of that,  LATENT can tolerate this loss of probability.  Consequently, UER allows changing the amount of randomization of the input bits by changing $\alpha$ while maintaining the same privacy budget. 

\label{probdef} .

\subsubsection{Generate a differentially private classification model using the  FC module}

After randomizing the merged binary strings, LATENT feeds the randomized binary strings into the FC module of the convolutional network. The FC module is then trained on the randomized binary strings to generate a differentially private ANN model (DPFC module). We improve the performance of the differentially private model using regularization, image augmentation, and hyperparameter tuning. 

\subsection{Algorithm for Generating a Differentially Private CNN}
Algorithm \ref{ranalgo} shows the steps of LATENT in producing a differentially private output. It provides the sequence of steps of applying differential privacy to the CNN architecture, as explained in Section \ref{convolearn} .

\begin{center}
    \scalebox{0.9}{
    \begin{minipage}{1.1\linewidth}
     \removelatexerror
      \begin{algorithm}[H]
    \caption{Differentially private model generation}
    \label{ranalgo}
            \KwIn{
            \begin{tabular}{l c l} 
            $\{x_1,\dots, x_j\}               $ & $\gets $ & examples\\
              $\varepsilon              $ & $\gets $ & privacy budget\\
             $n              $ & $\gets $ & number of bits for the  whole \\
             & & number of  the  binary\\
             & & representation\\
             $m              $ & $\gets $ & number of bits for the fraction\\
             & & of the binary representation\\
             $\alpha$ & $\gets$ & privacy budget coefficient
             \end{tabular}
             }
				
			\KwOut{
			\begin{tabular}{ l c l } 
				$DPCNN$ & $ \gets $ &  differentially private CNN model 
			\end{tabular}
			}
            Define the convolutional module (CNM) as explained in Section \ref{convolearn}\; 
            Declare, $l=(m+n+1)$\;
            Feed $\{x_1,\dots, x_j\}$ to the CNM and generate the sequence of 1-D feature arrays $\{d_1,\dots, d_j\}$\;
            Convert  each field ($x$) of $d_q$ (where, $q=1,\dots,j$) to binary using, $g(i)=\,{\Big(\left\lfloor 2^{-k}\, \abs x \right\rfloor\text{ mod }2\Big)_{k=-m}^{n}}\ \text{where, } i=k+m$\;
         
            Generate array $\{b_1,\dots, b_j\}$ of the merged binary arrays  for the elements in $\{d_1,\dots, d_j\}$\;
            Determine the length ($r$)  of a single element of $\{d_1,\dots, d_j\}$\;
			Calculate randomization probability according to Eq. \ref{twomodprob}\; 
            Randomize each element of $\{b_1,\dots, b_j\}$ using UER (refer Theorem \ref{uertheorem}) with probability $p$ to generate $\{pb_1, \dots, pb_j\}$\;

            Train the FC module of the CNN using $\{pb_1, \dots, pb_j\}$ \;
            Optimize the FC module using regularization, image augmentation and/or hyperparameter tuning\;
            
            Return the DPFC module\;  
   
      \end{algorithm}
    \end{minipage}%
}
\end{center}

\subsection{The LDP Settings for LATENT}

\begin{figure}[H]
    \centering
    \scalebox{0.21}{
    \includegraphics[width=1\textwidth, trim=0.5cm 0cm 0.5cm
     0cm]{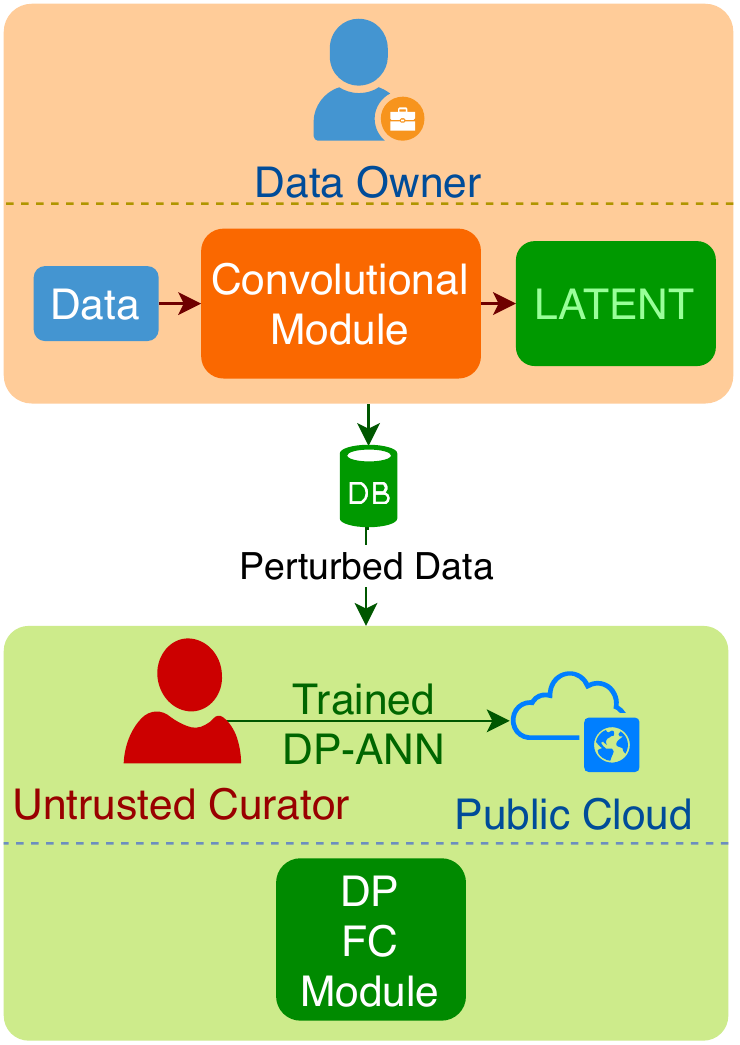}
     }
    \caption{LDP configurations of LATENT}
    \label{latentgdpldp}
\end{figure}

Since the randomization takes place after the convolutional module, we push the convolutional module and the LATENT module to the data owner's end as shown in Fig. \ref{latentgdpldp}.  The DPFC module is trained at the untrusted curator's end, and can be executed on a cloud computer or on any high-performance computing server. The model release will involve the release of only the trained DPFC module which can be used for testing by any third party. In the proposed setting, the convolutional module is not trained for features, leaving a minimum computational burden on a particular data owner.

The proposed component distribution of the CNN architecture, which moves the convolutional module to the data owner, produces additional privacy even before the randomization, as the output of the CNN module is a dimension-reduced 1-D vector. Additionally, in the big data context where millions of data owners communicate with the server, our CNN model distribution can provide additional flexibility and efficiency in data processing, leaving the FC module to train on the already dimension-reduced data. 

\begin{figure}[H]
	\centering
	\scalebox{0.43}{
	\includegraphics[width=1\textwidth, trim=0.5cm 0cm 0.5cm
	 0cm]{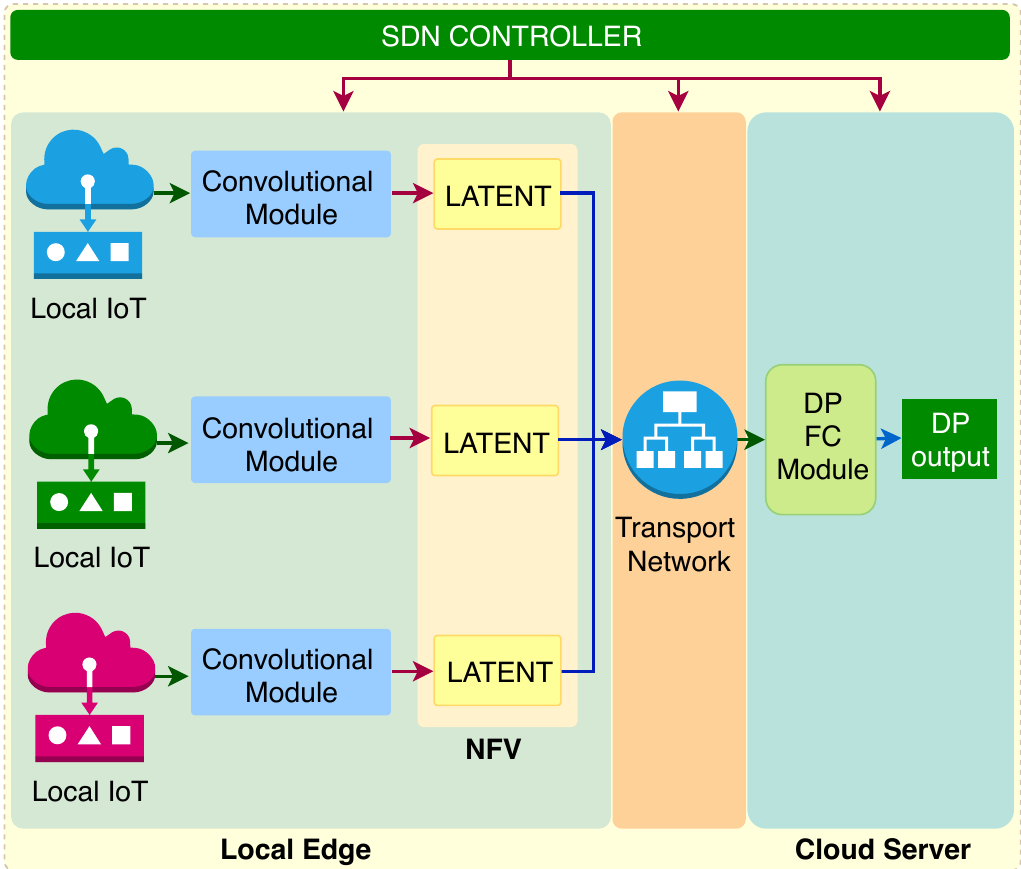}
	 }
	\caption{Integration of LATENT with SDN and NFV in edge-cloud interplay}
	\label{sdn_nfv}
\end{figure}

However, we can also push the whole DP CNN architecture to a single machine where we keep a central repository that is maintained by a trusted curator. Then we can apply the CNN with LATENT on the dataset at the trusted curator's end where the model will be released with the whole architecture of the CNN (Convolutional module (untrained)+LATENT+DPFC module).

\subsection{Integrating LATENT in the Amalgamation of SDN and  NFV in Edge-Cloud Interplay}

We can distribute the layers in LATENT (refer Fig. \ref{latentcnn}) in an SDN+NFV setting, as shown in Fig. \ref{sdn_nfv}. In this configuration, the first two layers of Fig. \ref{sdn_nfv} will reside at the local edge. First, the output from local IoT will go through the Convolutional module. The randomization layer will run as an NFV service which applies the randomization to the outputs from the convolutional modules. Consequently, the public transport layer will receive a randomized version of the input data and pass them on to the DPFC module in the cloud server, which produces a differentially private output. One or more SDN controllers will control the whole communication setup, as shown in Fig. \ref{sdn_nfv}.

\begin{figure}[H]
	\centering
	\scalebox{0.4}{
	\includegraphics[width=1\textwidth, trim=0.5cm 0cm 0.5cm
	 0cm]{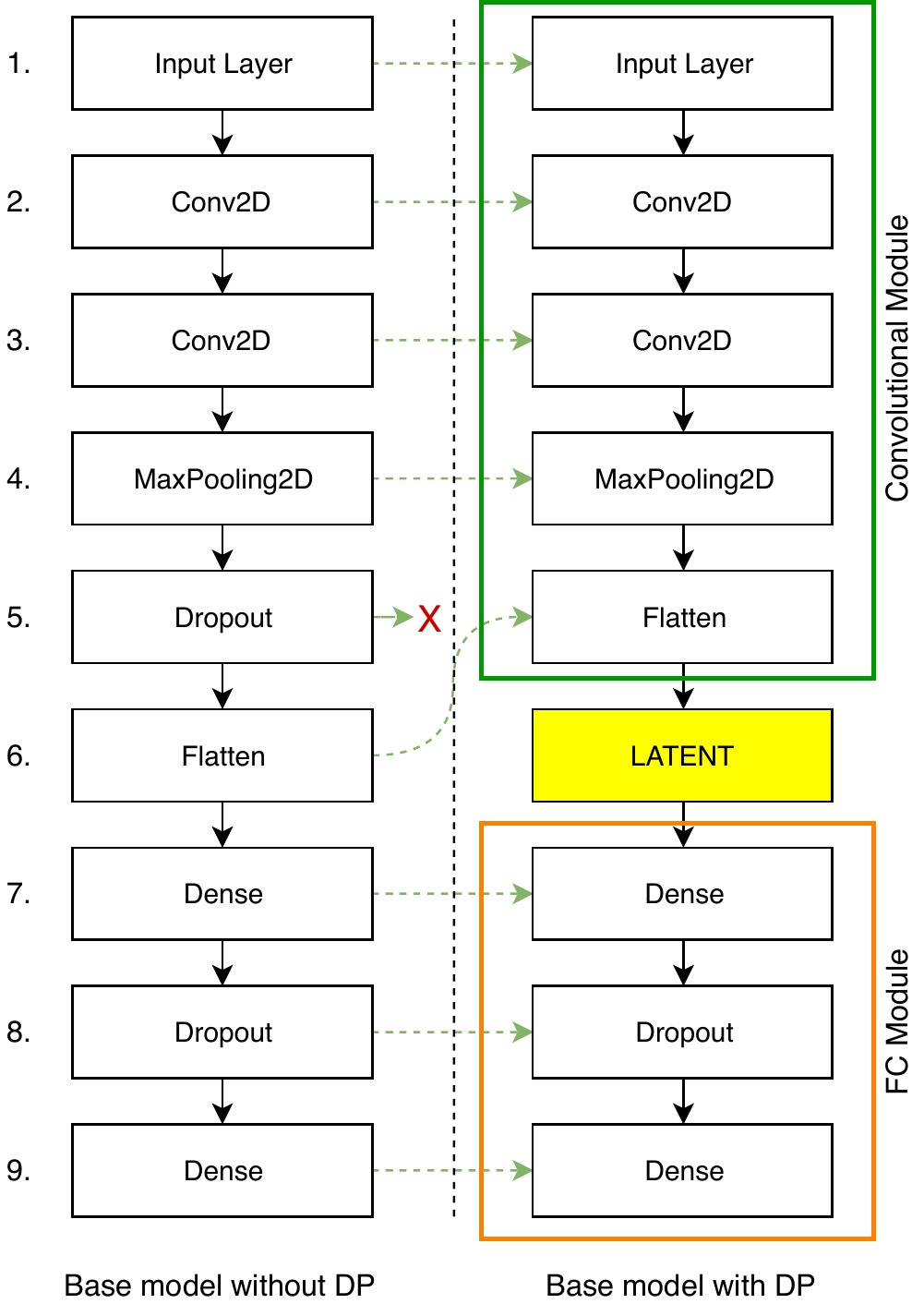}
	 }
	\caption{Architectural differences between the non-private (NPCNN) and differentially private (DPCNN) baseline models for the handwriting recognition dataset (MNIST dataset)}
	\label{ltmnst}
\end{figure}

\section{Results and Discussion}
\label{resdis}

In this section, we discuss the experiments, experimental configurations, and their results. We tested our method using the MNIST dataset~\cite{lecun1998gradient} and the CIFAR-10 dataset~\cite{abadi2016deep} which are considered to be the benchmark datasets to train and test deep learning (CNN) algorithms. We specifically selected MNIST  and CIFAR-10 for the experiments as they have been used in recent works on deep learning with differential privacy~\cite{shokri2015privacy,abadi2016deep}. MNIST is famous for generating good accuracy in deep learning, whereas CIFAR-10 is a complex dataset and is difficult for training. These complementary properties of MNIST and CIFAR-10 provide a balanced experimental setup to test the performance of a specific deep learning scenario.  We conducted all experiments on an HPC cluster (SUSE Linux Enterprise Server 12 SP3) with 112 Dual Xeon 14-core E5-2690 v4 Compute Nodes each with 256 GB of RAM,  FDR10 InfiniBand interconnect, and 4 NVidia Tesla P100 (SXM2). The computational complexity and the computational burden of LATENT on resource-constrained data owners was evaluated using a general purpose Intel Core i5 computer. A comprehensive specification of the corresponding computer is provided in Section \ref{compbr}.

\begin{figure}[H]
	\centering
	\scalebox{0.40}{
	\includegraphics[width=1\textwidth, trim=0.5cm 0cm 0.5cm
	 0cm]{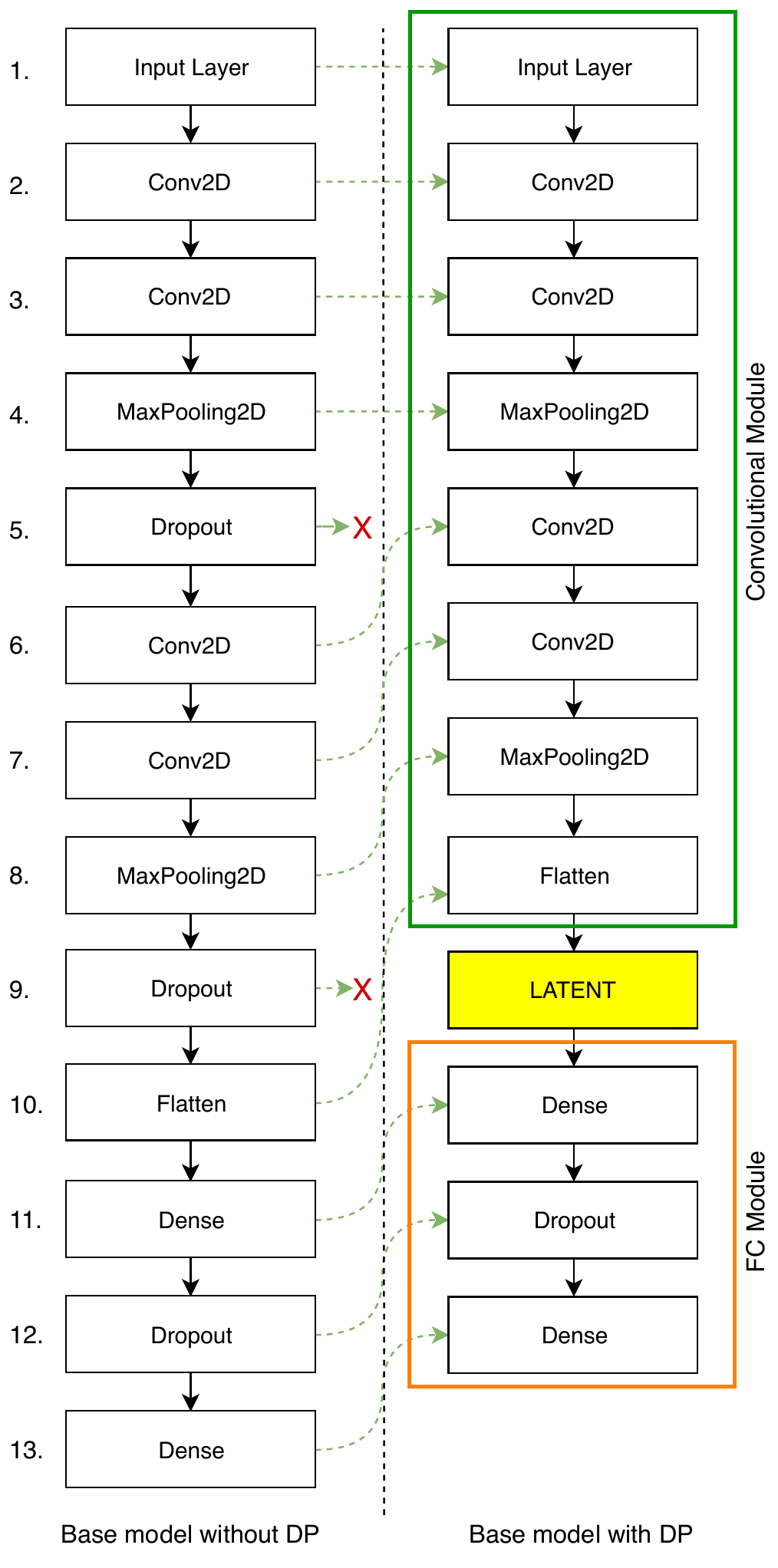}
	 }
	\caption{Architectural differences between the baseline models (NPCNN and DPCNN)  for image recognition of the CIFAR-10 dataset}
	\label{ltcifar}
\end{figure}

\subsection{Experimental Setup}

First, we created suitable baseline CNN models for each dataset. The baseline models include CNN without differential privacy (NPCNN) and a differentially private version of the same configuration (DPCNN). Fig. \ref{ltmnst} and Fig. \ref{ltcifar} show the baseline CNN architectures defined for the MNIST dataset and the CIFAR-10 datasets respectively. In the figures, the left-hand side models represent the NPCNN.  The right-hand side models are the differentially private versions (DPCNN) of the corresponding left-hand side models. First we tested the accuracy of the NPCNN models, then we tested the DPCNN models' performance relative to NPCNN models, and conducted hyperparameter tuning and image augmentation on the DPCNN models to improve performance. Finally, the results of the best DPCNN models were chosen to compare the results with other existing differentially private methods for deep learning.

\subsubsection{Datasets and CNN model information}
This section provides information about the datasets and the architectures of the corresponding CNN models used in the experiments. The architecture of a CNN needs to be custom configured, as the performance depends on the characteristics of the input dataset. The model quality of the trained ANN depends on the correct configuration of its network architecture~\cite{schmidhuber2015deep}.  As explained below, we declared suitable CNN architectures for the two datasets separately, because CIFAR-10 is a more complex dataset than MNIST.

\paragraph{MNIST}
The MNIST dataset is composed of 70,000 grayscale handwritten digits, where 60,000 examples are used for training, and 10,000 are used for testing. Each image has a resolution of  28x28. The digits have been size-normalized and centered in a fixed-size image~\cite{lecun1998gradient}. Fig. \ref{ltmnst} depicts the CNN network architecture used in the baseline models for the MNIST dataset. The figure shows the sequence of the layers of the network architectures of the baseline models. The network accepts 28 $\times$ 28 input images. The convolutional layer (layer 2) uses 32, 3 $\times$ 3 filters with stride 1 followed by a second convolutional layer (layer 3) which uses 64, 3 $\times$ 3 filters with stride 1. Both layer 2 and 3 use ReLU as the activation function. The output of layer 3 is subjected to a max pooling layer with 2$\times$2 max pools. Thus, the max pooling layer outputs a 12$\times$12$\times$64 tensor for each image. Next, the output of the max pooling layer is subjected to a dropout of 25\% (layer 5) and flattened (layer 6) to a 1-D vector of size 9216. The output of the flattening layer is fed into a fully connected layer (layer 7) with 128 neurons with ReLU activation function, followed by a dropout of 50\%. The output of the dropout layer is finally fed into a fully connected layer with 10 neurons which produces the final output of the CNN network. This model (NPCNN) achieves 99.25\% training and 98.16\% testing accuracies after 12 epochs of training with a batch size of 128 using the Adadelta optimization algorithm.

\paragraph{DPCNN for MNIST} The DPCNN has an additional layer: LATENT (layer number 6, colored in yellow) of randomization in between the convolutional module and the FC module. The green square represents the convolutional module, and the orange square represents the DPFC module. If the LATENT layer uses 10 bits to represent one element (one output of the flattening layer) coming from the fattening layer, the length of the randomized bit string generated by the LATENT layer is equal to 10 times the number of outputs of the flattening layer. 
In this case, the length of the randomized bit string will be equal to 9216 $\times$ 10 = 92160.  The green arrows are used to indicate that the same configuration of the corresponding layer is available in the DPCNN.  The red cross is used to indicate that the corresponding layer was omitted from the DPCNN. As shown in the figure, we do not use dropouts in the convolutional module of the DPCNN, as the convolutional module is not trained for the input features, which is explained in  Section \ref{convolearn}. In the DPCNN experiments with the MNIST dataset, we maintained a fixed size of 10 bits to represent each output of the flattening layer. The 10 bits are composed of 4 bits for the whole number, 5 bits for the fraction and 1 bit for the sign.

\begin{table}[H]
\caption{List of values applied for each hyperparameter in the test case generation process of hyperparameter tuning}
\centering
\resizebox{1\columnwidth}{!}{

\begin{tabular}{|l|l|l|}
\hline
\textbf{Hyperparameter} & \textbf{\begin{tabular}[c]{@{}l@{}}Hyperparameter settings\\ for MNIST\end{tabular}}   & \textbf{\begin{tabular}[c]{@{}l@{}}Hyperparameter settings\\ for CIFAR-10\end{tabular}} \\ \hline
percentage dropout      & \begin{tabular}[c]{@{}l@{}}layer 8 =\textgreater\\ \{20\%,  40\%,   \textcolor{red}{50\%}\}\end{tabular} & \begin{tabular}[c]{@{}l@{}}layer 11 =\textgreater \\ \{20\%, 40\%, \textcolor{red}{50\%}\}\end{tabular}  \\ \hline
batch size              & \{ \textcolor{red}{128}, 256, 512\}                                                                       & \{400, \textcolor{red}{500}, 600\}                                                                          \\ \hline
activation function                                                       & \begin{tabular}[c]{@{}l@{}}layer 7=\textgreater \\ \{\textcolor{red}{relu}, tanh, sigmoid\}\end{tabular}      & \begin{tabular}[c]{@{}l@{}}layer 10=\textgreater \\ \{\textcolor{red}{relu}, tanh, sigmoid\}\end{tabular}       \\ \hline

the number of neurons   & \begin{tabular}[c]{@{}l@{}}layer 7=\textgreater \\ \{64, \textcolor{red}{128}, 256, 512\}\end{tabular}     & \begin{tabular}[c]{@{}l@{}}layer 10=\textgreater\\  \{256, \textcolor{red}{512}, 1024\}\end{tabular}      \\ \hline
number of epochs        & \{50,  \textcolor{red}{100}, 150\}                                                                       & \{100, \textcolor{red}{200}, 300\}                                                                       \\ \hline
optimizer                                                                 & \{SGD, \textcolor{red}{Adadelta}, Adam\}                                                        & \{SGD, Adadelta, \textcolor{red}{Adam}\}                                                             \\ \hline
\end{tabular}
}
\label{hptprocess}
\end{table} 

\paragraph{CIFAR-10}
The CIFAR-10 dataset consists of 60000 color images and 10 classes (airplane, automobile, bird, cat, deer, dog, frog, horse, ship, and truck), with 6000 images per class. There are 50000 training images and 10000 testing images.  Each image has a resolution of  32$\times$32~\cite{abadi2016deep}. Fig. \ref{ltcifar} depicts the CNN architecture used in the baseline models for the CIFAR-10 dataset. As CIFAR-10 is a complex dataset compared to MNIST, for CIFAR-10 we considered a more complex CNN architecture which involves more layers and more neurons. The network accepts 32 $\times$ 32, 3 channel input images. The convolutional layer (layer 2) uses 32, 3 $\times$ 3 filters with stride 1 followed by a second convolutional layer (layer 3) which uses 32, 3 $\times$ 3 filters with stride 1. Both layers 2 and 3 use ReLU as the activation function.  The output of layer 3 is subjected to a max pooling layer with 2$\times$2 max pools. The output of layer 4 is fed to a dropout layer (layer 5) with 25\% dropout. The output of layer 5 was subjected to two other convolutional layers (layer 6 and 7) which use 64, 3$\times$3 filters. The output of layer 6 was next introduced with a max pooling layer (layer 8) with 2$\times$2 max pools.  Thus, the max pooling layer outputs a 6$\times$6$\times$64 tensor for each image. Next, the output of the max pooling layer is subjected to a dropout of 25\% (layer 9), and flattened (layer 10) to a 1-D vector of size 2304. The output of the flattening layer is fed into a fully connected layer (layer 11) with 512 neurons with ReLU activation function, followed by a dropout of 50\%. The output of the dropout layer is finally fed into a fully connected layer with 10 neurons which produces the final output of the CNN network. This model (NPCNN) achieves 73.32\% training and  78.75\% testing accuracies after 100 epochs of training for a batch size of 32 using the  Adadelta optimization algorithm.

\paragraph{DPCNN for CIFAR-10} The right-hand side figure of Fig. \ref{ltcifar} represents the DPCNN for the CIFAR-10 dataset. The DPCNN model creation follows the same approach explained under the model creation process for the MNIST dataset. However, we increased the resolution of the images to 56$\times$56 to enhance the features. As a result of that, the input layer (layer 1) was customized to accept 56$\times$56, 3 channel input images.  We do not use dropouts in the convolutional module for reasons explained in Section \ref{convolearn}.  In the experiments with the DPCNN model for the CIFAR-10 dataset, we maintained a fixed size of 10 bits to represent each output of the flattening layer. The 10 bits are composed of 2 bits for the whole number, 7 bits for the fraction and 1 bit for the sign.

\subsubsection{Hyperparameter tuning and regularization}
\label{hytrain}
As explained in Section \ref{convolearn},  we conducted the training only on the FC module.  Therefore, we apply hyperparameter tuning and regularization only on the FC module.  Given the number of possible values for each hyperparameter, the number of test cases can become large, and it may entail a substantial computational cost with exponential time.  
Therefore, it can be imperative to use insights such as used by \cite{abadi2016deep} to minimize the number of hyperparameter settings that need to be tested. Since LATENT does not change the internal parameters of the FC module, the architectural modifications necessary can be thought of as an independent procedure that can be common in any ANN training process. Consequently, we tested different combinations for percentage dropouts, batch sizes, activation functions, number of neurons, optimizers, number of epochs, for the hyperparameter tuning process as given in Table \ref{hptprocess}.

The values used for the hyperparameters during the  HPT processes with each dataset (MNIST and CIFAR-10) are given in Table \ref{hptprocess}. We preferred higher number of neurons and epochs for CIFAR-10 due to its complexity compared to MNIST. During the hyperparameter tuning process, the probability of preserving the true value of an original bit ($p$)  was set to 1,  which corresponds to the nonprivate state of LATENT. When p=1, the binary feature vectors will not be randomized, and the model will result in greater accuracy than when p<1. Due to the enlarged feature space compared to the conventional 1-D output of the flattening layer, the input features will have more representative properties. These properties will allow the proposed architecture of CNN to generate better accuracy than that of the original CNN architecture without LATENT. We generated the test cases using combinations of parameter values. We applied k-fold cross-validation (k=10) on each test case to derive a fair set of accuracy results. Since the search space is large, we divided the list of test cases into nine groups based on the combinations of the optimizers and the activation function. The best parameter values returned for each dataset are colored in red in Table \ref{hptprocess}. Next, we used the best parameters returned by the tuning process to produce differentially private models (for MNIST and CIFAR-10) with the best performance. We used the final DPCNN models of the hyperparameter tuning process to carry out further experiments and analyses.

\begin{figure}[H]
	\centering
	\subfloat[FC module training with $\alpha =5$]{\includegraphics[width=0.23\textwidth, trim=0cm 0cm 0cm 0cm]{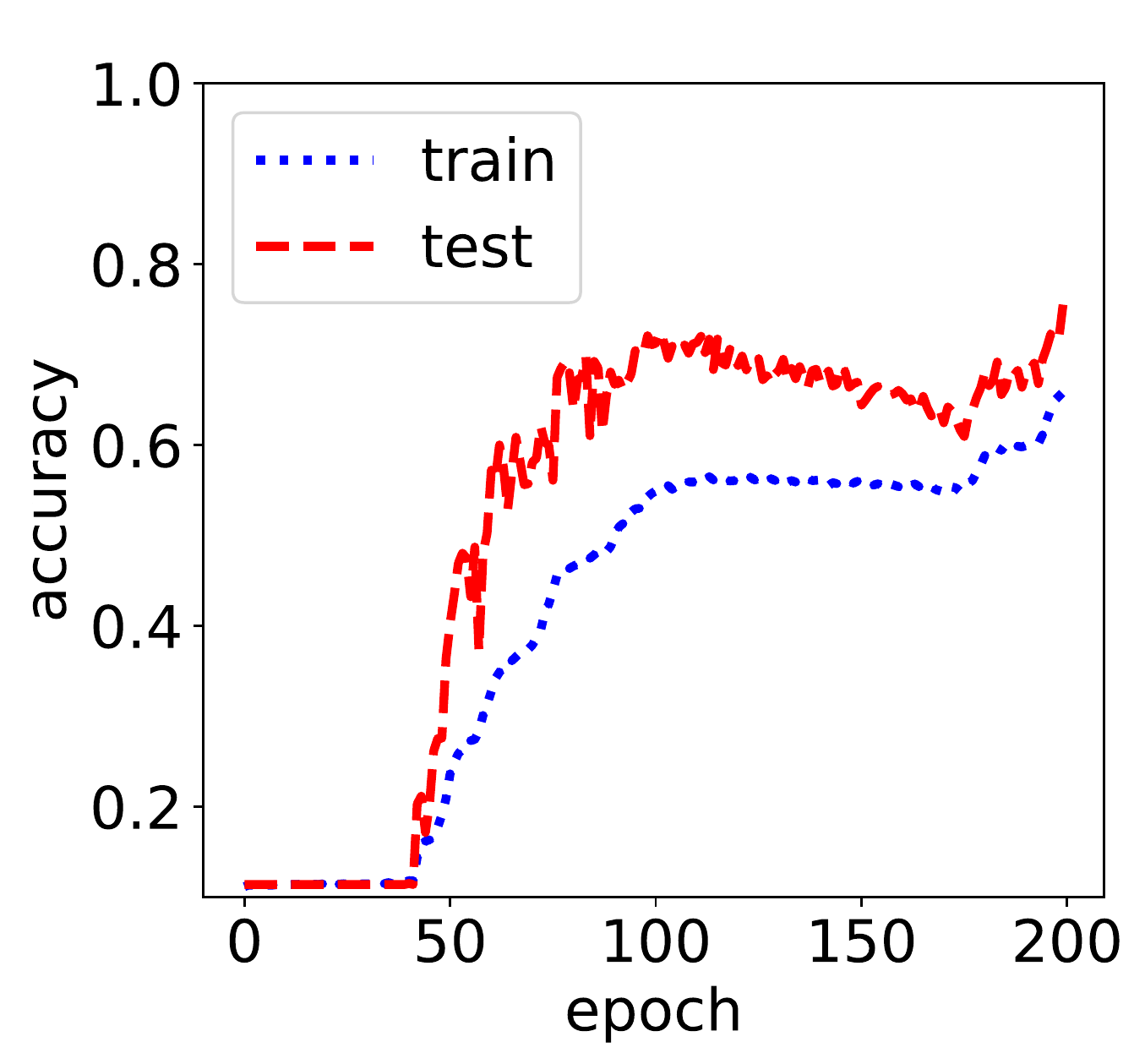}\label{accuplotMNIST1}}
	\hfill
	\subfloat[FC module training with $\alpha =6$]{\includegraphics[width=0.23\textwidth, trim=0.3cm 0cm 0cm 0cm]{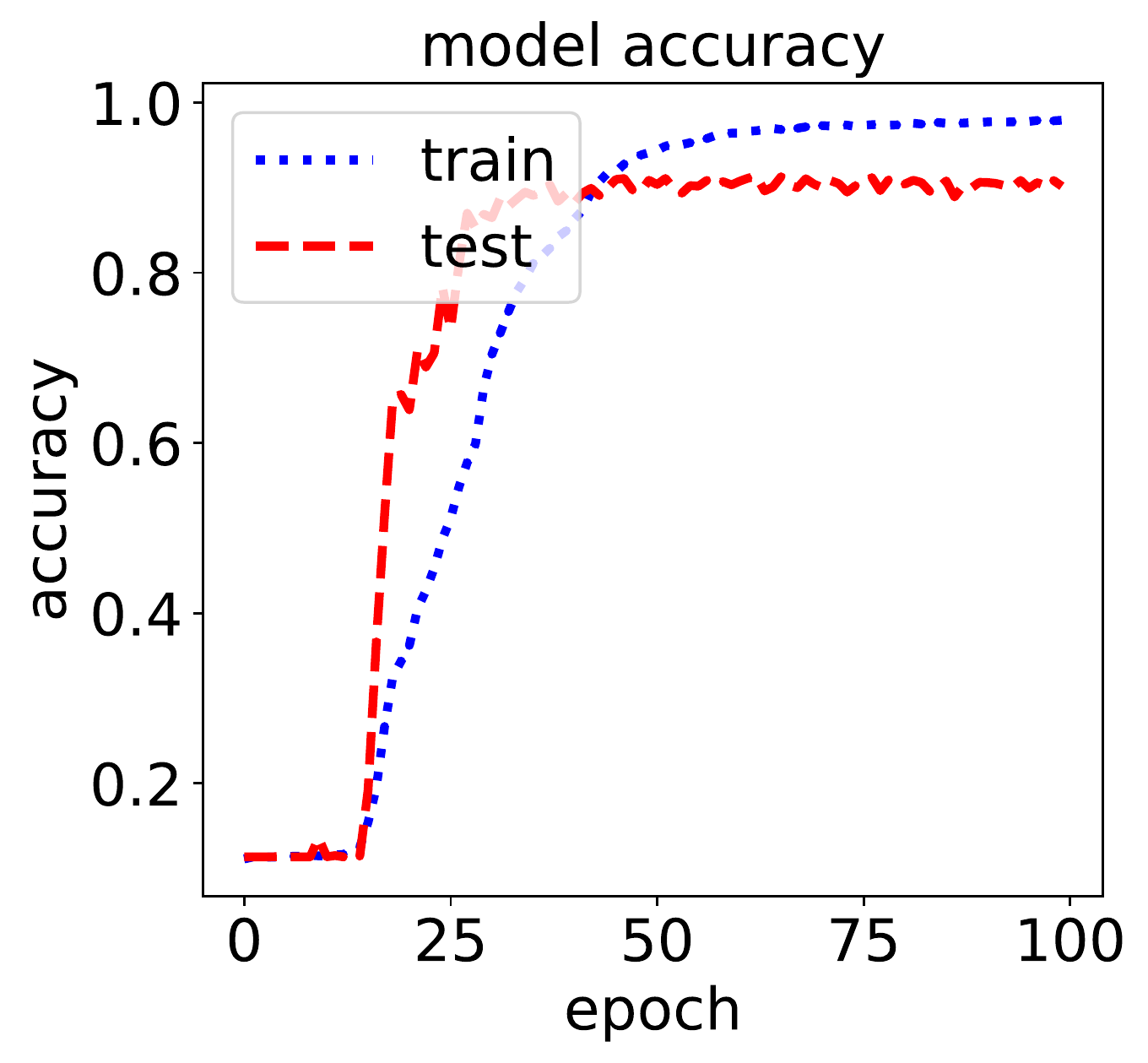}\label{accuplotMNIST2}}
	\\
	\subfloat[FC module training with $\alpha =7$]{\includegraphics[width=0.23\textwidth, trim=0cm 0cm 0cm 0cm]{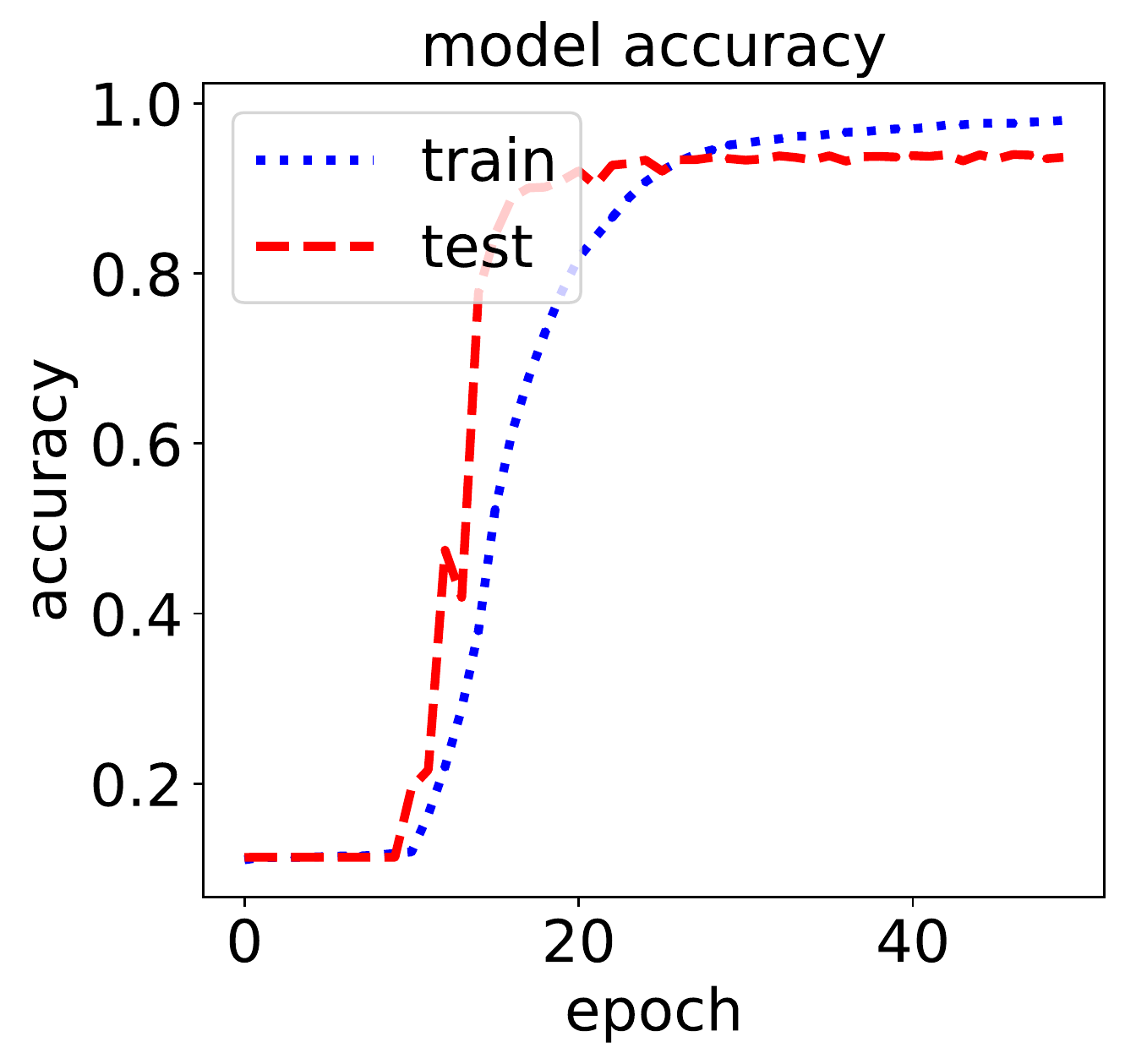}\label{accuplotMNIST3}}
	\hfill
	\subfloat[FC module training with $\alpha =8$]{\includegraphics[width=0.23\textwidth, trim=0.3cm 0cm 0cm 0cm]{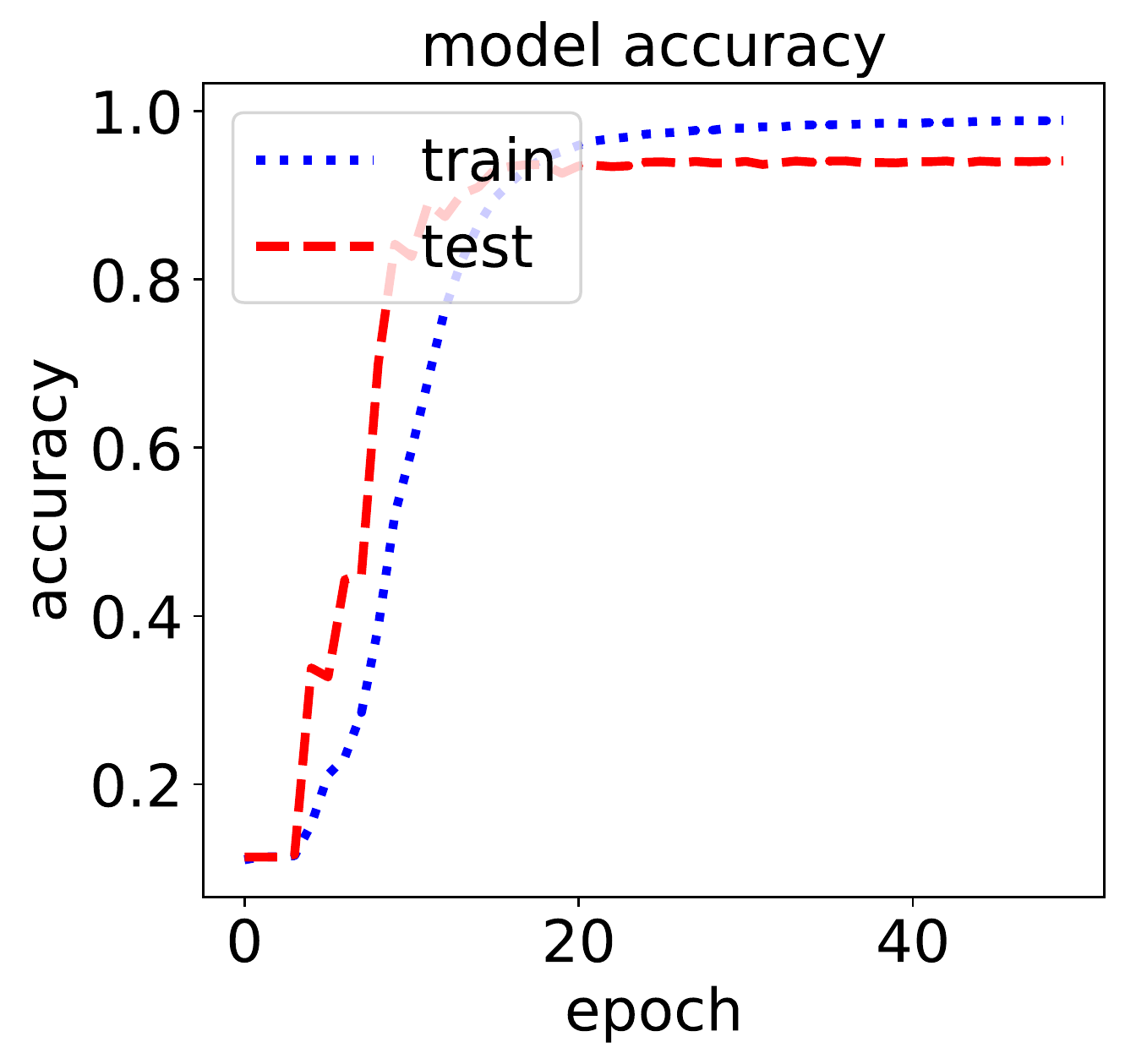}\label{accuplotMNIST4}}
 	\caption{The model convergence of the FC module for the MNIST dataset (under $\varepsilon=0.5$ and the chosen hyper-parameters which are red-colored in Table \ref{hptprocess} for MNIST) under different $\alpha$ values (randomization levels)}
 	\label{accuplotMNIST}
 	\medskip
    \small
\end{figure}

\begin{figure}[H]
	\centering
	\scalebox{0.33}{
	\includegraphics[width=1\textwidth, trim=0.5cm 0cm 0.5cm
	 0cm]{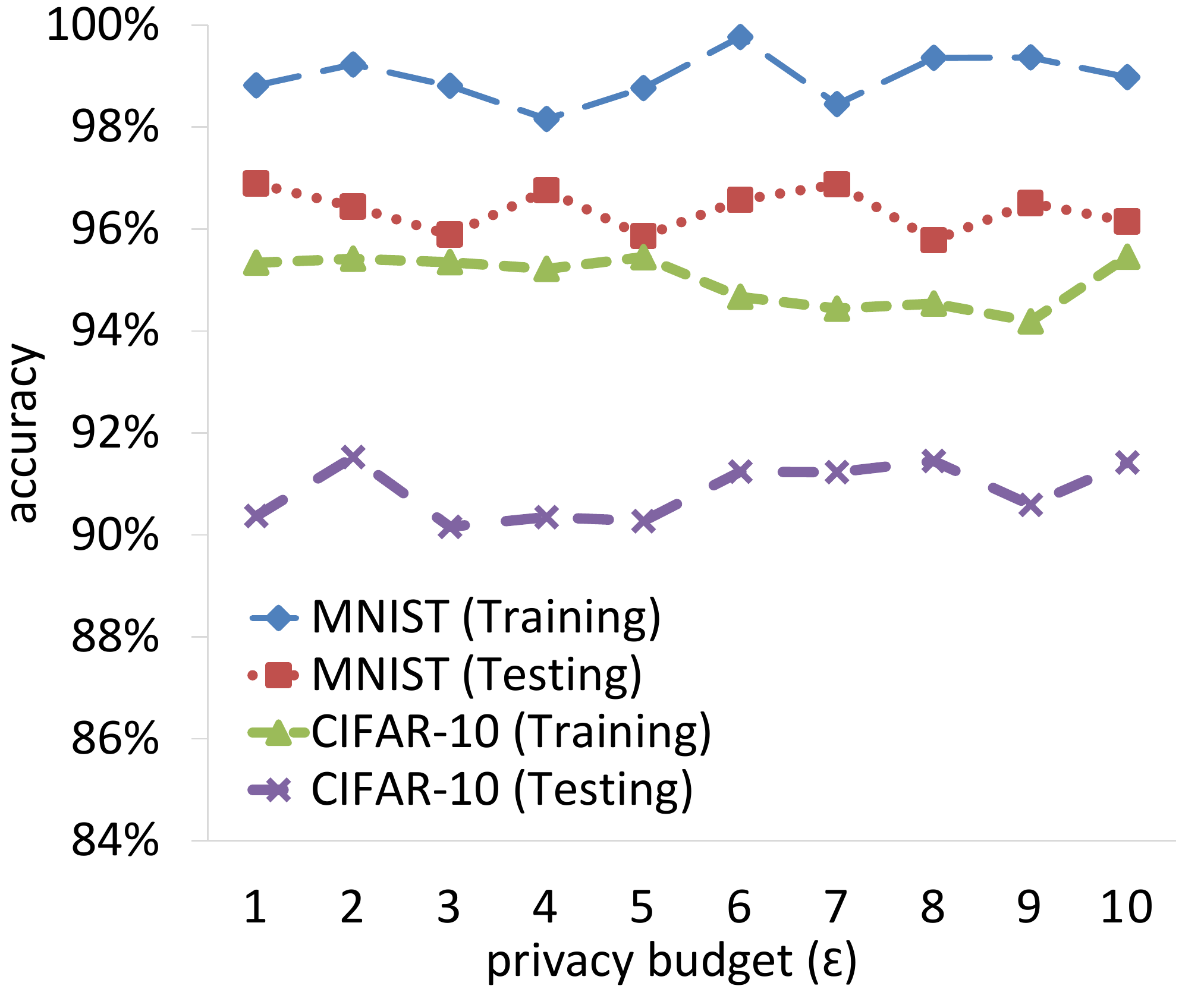}
	 }
	\caption{Change of accuracy of LATENT against $\varepsilon$}
	\label{accepsilon}
\end{figure}

\subsubsection{Image augmentation to improve robustness of the DPCNN trained using CIFAR-10 }

Although we could improve the accuracy of the DPCNN for CIFAR-10 using hyperparameter tuning, the model was still not performing well and tended to overfit. To improve the model robustness, we applied image augmentation and generated 150,000 additional augmented images using the 50,000 training input images. Each augmented image was generated by applying a random horizontal shift of a 0.1 fraction of the total width, a random vertical shift of a 0.1 fraction of the total height, a random rotation of 10 degrees, and a random horizontal flip, on the original input images. After introducing the new augmented images, the DPFC module stopped overfitting and started generating a training accuracy of around 98\% and a testing accuracy of about 95\% consistently for repeated attempts (under a randomization probability of 1).

\begin{table*}[t]
 \caption{Accuracy comparison of the results of LATENT against the existing methods}
 \centering
\resizebox{1.4\columnwidth}{!}{
\begin{tabular}{|l|l|l|l|l|l|l|l|}
\hline
\multicolumn{2}{|c|}{\textbf{Dataset}} & \multicolumn{1}{c|}{\textbf{NPCNN}} & \multicolumn{1}{c|}{\textbf{{[}SS15{]~\cite{shokri2015privacy}}}}                                                        & \multicolumn{2}{c|}{\textbf{{[}ACG+16{]~\cite{abadi2016deep}}}}                                                                                                          & \multicolumn{2}{c|}{\textbf{LATENT}} \\ \hline
\multicolumn{2}{|l|}{}                 & \begin{tabular}[c]{@{}l@{}}accuracy of \\ the model \\without privacy\end{tabular}                          & \begin{tabular}[c]{@{}l@{}}$\varepsilon$ is large\\ as it is reported\\ per parameter\end{tabular} & \begin{tabular}[c]{@{}l@{}}$\varepsilon=2$\\ $\delta=10^{-5}$\end{tabular} & \begin{tabular}[c]{@{}l@{}}$\varepsilon=0.5$\\ $\delta=10^{-5}$\end{tabular} & $\varepsilon = 2$ & $\varepsilon = 0.5$   \\ \hline
\multirow{2}{*}{MNIST}     & Training  & 99.25\%                             &  N/A                                                                                           & $\sim$95\%                                                              & $\sim$89\%                                                                & 98.46\%          & 98.97\%           \\ \cline{2-8} 
                           & Testing   & 98.16\%                             & 98\%                                                                                            & 95\%                                                                    & 90\%                                                                      & 95.67\%          & 96.37\%           \\ \hline
\multirow{2}{*}{CIFAR-10}  & Training  & 73.32\%                             & N/A                                                                                             & $\sim$68\%                                                              & N/A                                                                       & 95.62\%          & 95.01\%           \\ \cline{2-8} 
                           & Testing   & 78.75\%                             & N/A                                                                                             & 67\%                                                                    & N/A                                                                       & 91.34\%          & 91.47\%           \\ \hline
\end{tabular}
    }
  \label{accomp}%
\end{table*}%

\subsubsection{Selection of $\varepsilon$}

As we discussed in Section \ref{convolearn}, the probability of randomization can be given by Equation \eqref{mergedrandprob}. When $l=10$ for the DPCNN architecture defined for the MNIST dataset (depicted in Fig. \ref{ltmnst}), $p=\frac{e^{\varepsilon/92160}}{1+e^{\varepsilon/92160}}$ as the sensitivity of the DP mechanism is 92160. For the DPCNN defined for the CIFAR-10 dataset (depicted in Fig. \ref{ltcifar}) also, $p=\frac{e^{\varepsilon/92160}}{1+e^{\varepsilon/92160}}$, since the sensitivity is 92160 due to the increased resolution of the input images from 32$\times$32 to 56$\times$56. Because of the large sensitivity, the effect of $\varepsilon$ in generating the randomization probabilities is low. When $\alpha$ is maintained at 1, the probability of randomization ($p$) lies around $0.5$ for the acceptable values of $\varepsilon$ (less than 10) as the sensitivity of the processes for both models are large (refer Eq. \ref{twomodprob}).   Hence, we used $\varepsilon=0.5$ to generate the results in the experiments. We maintained $\alpha$ at 7 (unless mentioned otherwise), which was the lowest $\alpha$ value that generated reliable convergence of the models.

Figure  \ref{accuplotMNIST} shows the model convergence against different $\alpha$ values  against the number of epochs during the training process of the FC module on the MNIST dataset when $\varepsilon=0.5$. The FC module converges at different number of epochs against the different levels of randomizations introduced by the various values of $\alpha$. The converged models provide excellent training and testing  accuracies as shown in the plots of Figure  \ref{accuplotMNIST}. Clarity of the MNIST dataset and the availability of a large feature space generated by LATENT  allow the model to produce excellent accuracies: around 98-99\% for training and around 95-96\% for testing even at very low $\varepsilon$ values such as 0.5 and $\alpha>5$ under the chosen hyper-parameters which are red-colored in Table \ref{hptprocess}.

\begin{figure}[H]
	\centering
	\scalebox{0.3}{
	\includegraphics[width=1\textwidth, trim=0.5cm 0cm 0.5cm
	 0cm]{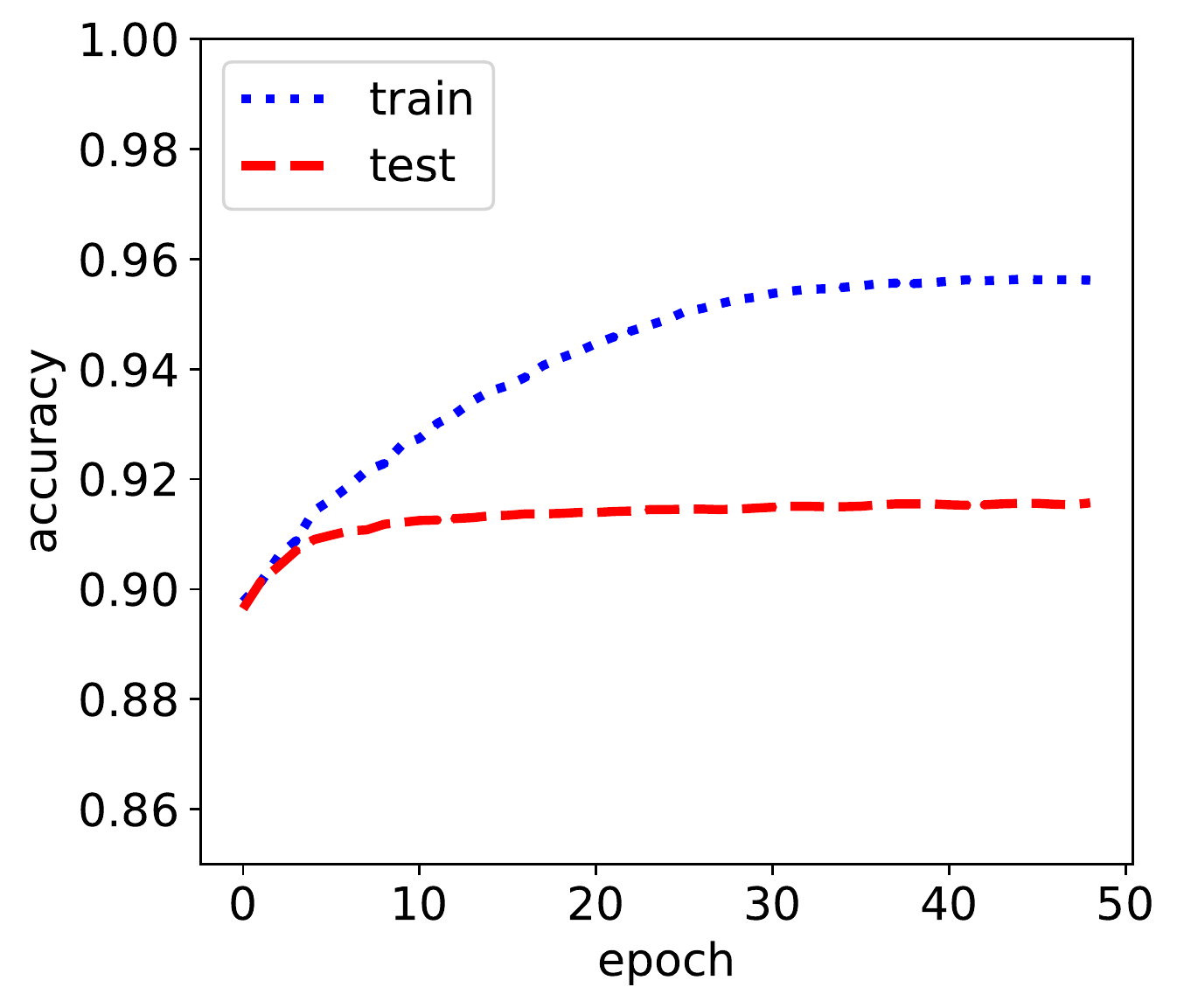}
	 }
	\caption{The change of accuracy vs. the number of epochs during the training of the DPFC module for the CIFAR-10 dataset (under $\varepsilon=0.5$ and the chosen hyper-parameters which are red-colored in Table \ref{hptprocess} for CIFAR-10)}
	\label{prbep}
\end{figure}

Fig. \ref{prbep} shows the change of accuracy against the number of epochs during the training process of the FC module of the CIFAR-10 dataset when $\varepsilon=0.5$. After increasing the input image resolution and applying image augmentation to generate 150,000 new images under the best-chosen hyper-parameters (red-colored in Table \ref{hptprocess}), the trained model returned around 95-96\% training accuracy and about 90-91\% testing accuracy after 50 epochs. The significant feature space generated by LATENT, and the large input space created by image augmentation allow the final model to produce the corresponding excellent accuracies with high robustness of the DPFC model.

Fig. \ref{accepsilon} shows the change of accuracy against $\varepsilon$ values. As the figure depicts, accuracy is almost constant although $\varepsilon$ is changed. Recall that the probability of randomization is loosely affected by small values of $\varepsilon$ (when $\alpha$ is constant) due to the high sensitivity values. When $\alpha$ is kept constant (=7), LATENT maintains a uniform level of randomization on each dataset under each case of $\varepsilon$ depicted in Fig. \ref{accepsilon} and produces similar accuracy for smaller values of $\varepsilon$ ($<10$).

\subsection{Comparison of the Results of LATENT Against the Existing Approaches}

We compare our results with two other existing differentially private mechanisms for deep learning, as shown in Table \ref{accomp}. Although [SS15]~\cite{shokri2015privacy} provides good accuracy; $\varepsilon$ is presented as a parameter of the model. It can accumulate a large, unacceptable $\varepsilon$ value at the end of model generation as there can be more than 1000 model parameters. For $\varepsilon=2$ and $\delta=10^{-5}$ the [ACG+16]~\cite{abadi2016deep} method provides good accuracy, yet the additive bound of $\delta$ can become unreliable when the method is used for much larger datasets. [ACG+16] has failed to generate acceptable accuracy for the CIFAR-10 dataset in an extreme case like $\varepsilon=0.5$. Our method provides much better accuracy for an extremely small privacy budget, such as $\varepsilon=0.5$. Also, the unavailability of additive bound ensures that our method has a low privacy leak when substantially large input datasets are presented to the method. Both [SS15] and [ACG+16] are based on global differential privacy. Therefore, the availability of a trusted party is unavoidable.  For a real-world scenario, often we don't have any trusted party. LATENT can be a much better solution in such cases, as it works with both untrusted and trusted curators. Table \ref{compmethod} sums up the advantages of LATENT over the existing GDP methods for deep learning.

\subsection{Computational Complexity and Computational Burden on Generic Users}
In  this section, we investigate the computational complexity of the proposed algorithm and its burden on a resource constrained computer. For these two experiments, we used a MacBook Pro (macOS Mojave, 13-inch, 2017) personnel computer (PC) with Intel Core i5  CPU (2.3 GHz), 8 GB RAM and 1536MB GPU (Intel Iris Plus Graphics), with the assumption that a general purpose computer is enough to work as an aggregator in a multi-sensor setup.

\begin{figure}[H]
	\centering
	\subfloat[Time consumption (in seconds) for binary encoding.]{\includegraphics[width=0.23\textwidth, trim=0cm 0cm 0cm 0cm]{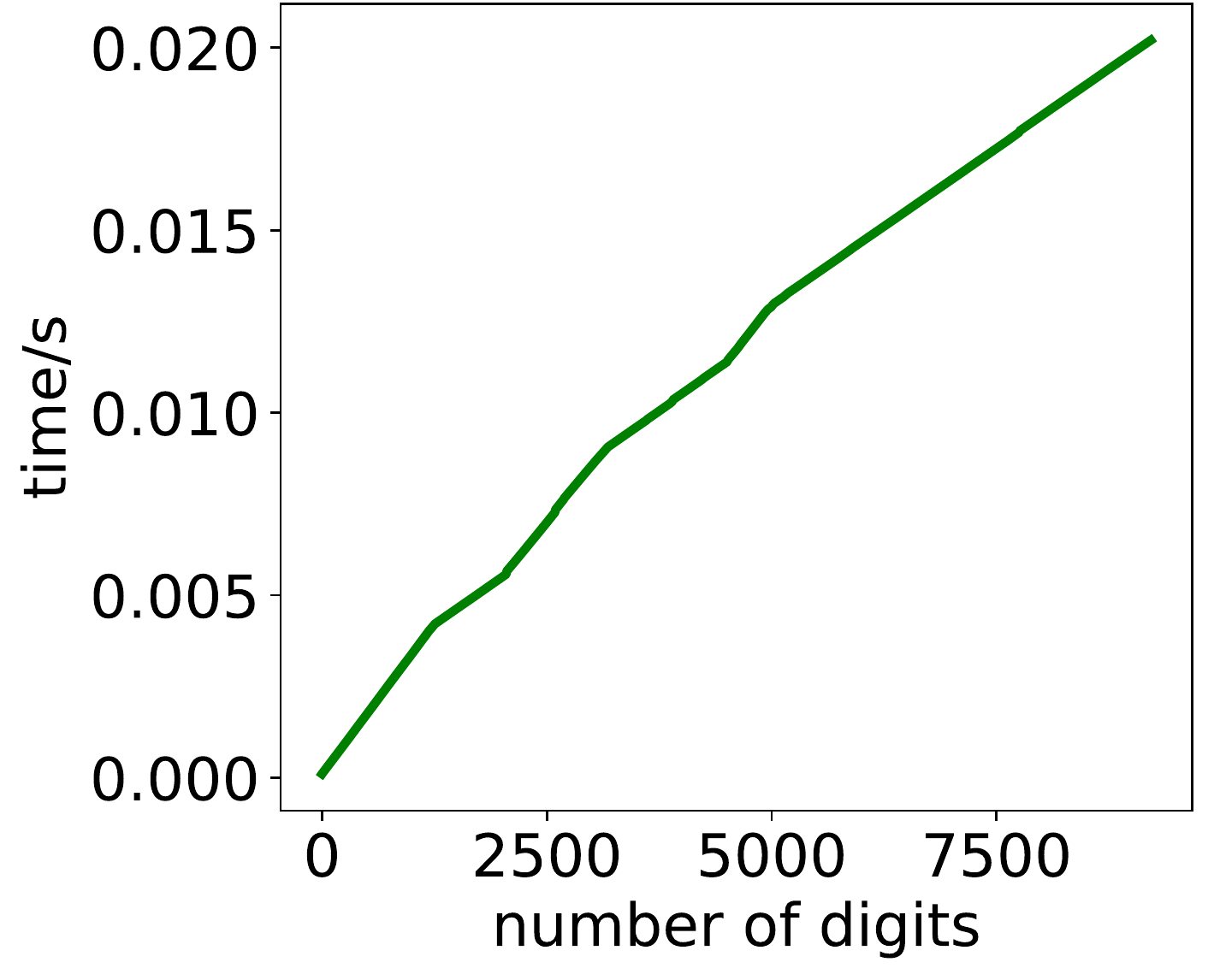}\label{bincomplex}}
	\hfill
	\subfloat[Time consumption (in seconds) for  randomization]{\includegraphics[width=0.23\textwidth, trim=0.3cm 0cm 0cm 0cm]{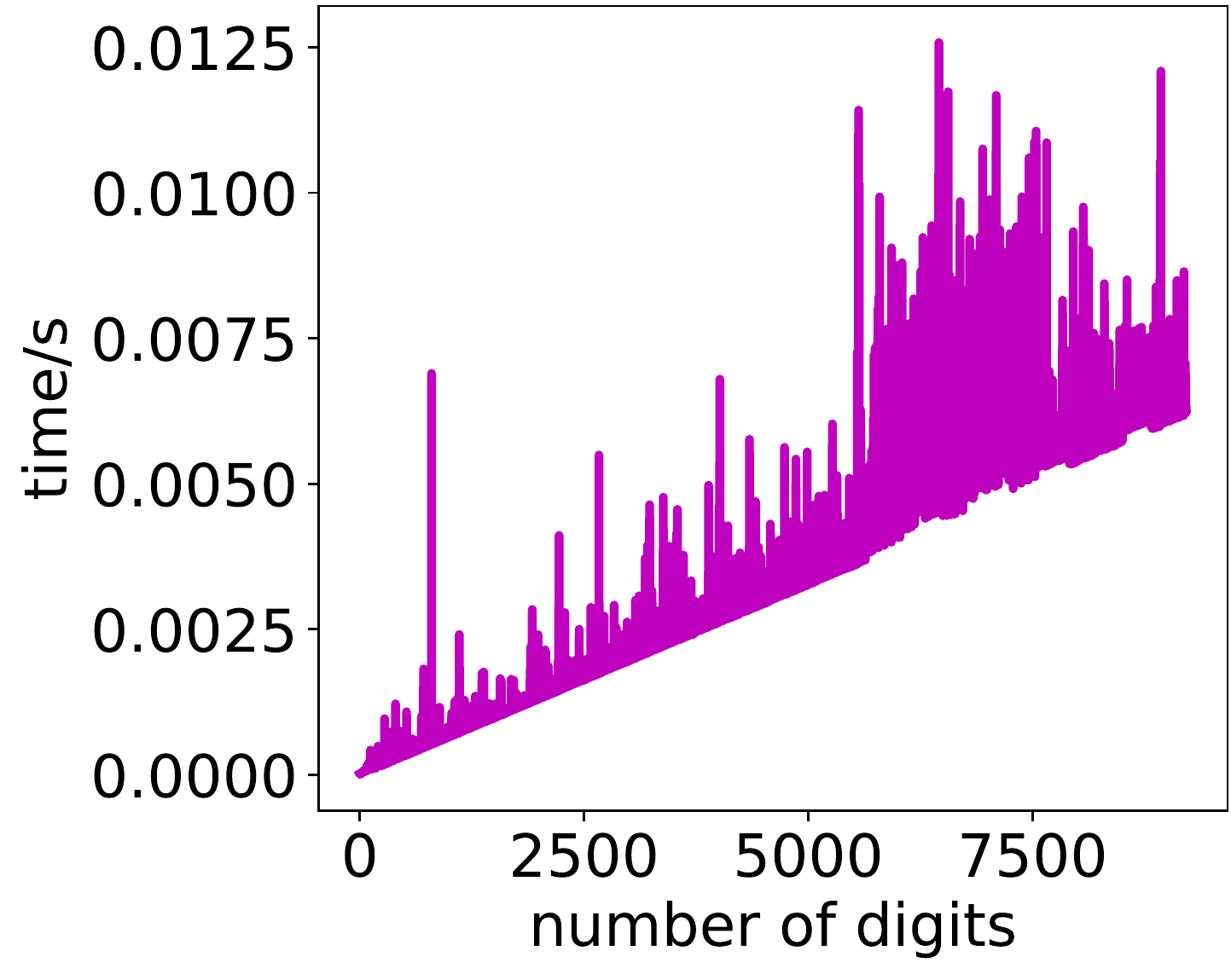}\label{rancomplex}}
 	\caption{Time consumption analysis for the binary encoding and randomization steps of LATENT against the number of digits generated by the convolutional module.}
 	\label{timecomplexityplt}
 	\medskip
    \small
\end{figure}

LATENT involves two main steps that govern its computation complexity: (1) binary conversion and (2) randomization of the convolutional module output. Each digit in the convolutional module output is subjected to binary conversion according to Eq. \ref{intbinary}. The parameters of the binary conversion (m and n, refer to Eq. \ref{intbinary}) are fixed at the beginning of Algorithm \ref{ranalgo}. Therefore, the computational complexity of the binary conversion step depends on the number of digits ($l$) in the convolutional output. The computational complexity of step 1 (binary conversion) can be derived as $O(l)$, i.e. is linear. This can be emperically proven by Fig. \ref{bincomplex}. The length of the binary string is determined by the values intialized to $m$ and $n$. An $l$ sized vector will produce a bit string of $l\times(m+n+1)$ (refer to Section \ref{binconversion}). Given that the randomization of each bit is independent, the complexity of the randomization step is also governed by $l$, which introduces a computational complexity of $O(l)$ as represented by Fig. \ref{rancomplex}.

\label{compbr}

In the proposed modular decomposition of the CNN architecture, the convolutional module and the LATENT module run on the data owner's machine. We need to make sure that the convolutional module and LATENT operations do not impose a substantial computational burden on the resource-constrained data owners. In order to check this,  we measured the time consumption for the perturbation of a single record of MNIST and CIFAR-10 datasets separately.  It took an average time of 0.1655 seconds to perturb a single record of MNIST dataset while consuming 0.0374 seconds to perturb a single record of CIFAR-10 dataset. This indicates that a general purpose computer with moderate specification will suffice for generating the randomized data.  We can conclude that the convolutional module and the randomization module of LATENT are feasible to be implemented in any IoT setting or equivalent in the cloud environment.

\begin{table}[]
\caption{Advantages of LATENT over the existing GDP methods for deep learning}
\label{compmethod}
\resizebox{1\columnwidth}{!}{
\begin{tabular}{@{}|l|l|@{}}

\hline
\multicolumn{1}{|c|}{\textbf{GDP methods}}                                                                                          & \multicolumn{1}{c|}{\textbf{LATENT}}                                                                                                                     \\ \hline
Always needs a trusted curator.                                                                                                                                                                                                                                                 & \begin{tabular}[c]{@{}l@{}}LATENT  can be used for \\ both trusted \\  and untrusted settings.\end{tabular}                                                                                                                                                                                             \\ \hline
\begin{tabular}[c]{@{}l@{}}For machine learning with   \\cloud computing, original \\ data needs to be uploaded\\ to the  server considering\\ the server  is trustable.  \\However, the servers cannot\\ always be trusted.\end{tabular}                                                                                                                                                                                                                                                         & \begin{tabular}[c]{@{}l@{}}LATENT randomizes data \\before uploading \\them to the server\\ in case the server is not \\ trustable.\end{tabular}                                                                                                                                                                                \\ \hline
\begin{tabular}[c]{@{}l@{}}A higher privacy loss (a larger  \\ privacy budg
et - $\varepsilon$) needs to be \\ allocated to obtain a better utility.\end{tabular}                                                                                                                                                                                                                                                                                                     & \begin{tabular}[c]{@{}l@{}}LATENT provides excellent \\ utility in terms of\\  classification accuracy\\ (more than 90\%)  \\even under an extreme \\ level of randomization \\ ($\varepsilon=0.5$).\end{tabular} \\ \hline

\begin{tabular}[c]{@{}l@{}}GDP runs either in client\\ side or a server side. \\ The distrust of the server\\ might prevent the   algorithm\\ being run on a server.  However,\\  deep learning algorithms \\tend to be complex and can\\   be complex for a general\\  purpose personal   computer, and\\ privacy preservation  techniques\\ often add more complexity. This \\ feature reduces  the practicality\\  of GDP algorithms for\\ deep learning.\\ \end{tabular} & 

\begin{tabular}[c]{@{}l@{}}As LATENT is an LDP \\ algorithm, it doesn't\\ have an obligation \\ to have a trusted curator.\\ As the proposed architecture \\ is already a distributed\\ version which utilizes\\ the computational power of \\data owners and the servers,\\ LATENT is more practical \\compared to GDP approaches.  \end{tabular}                                                                                                                                         \\ \hline
\end{tabular}
}
\end{table}

\section{Related Work}
\label{relwork}

Privacy-preserving data mining (PPDM) provides the capability of using data mining techniques without disclosing private information of the participating entities in the underlying database. However, the main challenge in PPDM is countering the capabilities of skilled adversaries~\cite{yang2017efficient, vatsalan2017privacy}. To overcome this challenge, PPDM uses data modification (data perturbation)~\cite{chen2005random, chen2011geometric}  and encryption~\cite{kerschbaum2017searchable} techniques. Data encryption-based approaches provide good security and accuracy. However, cryptographic methods often suffer from high computational complexity, which makes encryption unsuitable for large-scale data mining~\cite{gai2016privacy}. Compared to encryption, data perturbation utilizes lower computational complexity, which makes it effective for big data mining~\cite{xu2014building}.  Noise addition,  geometric transformation, randomization, data condensation, hybrid perturbation (i.e. using several perturbation techniques together) are some examples of data perturbation techniques~\cite{chamikaraprocal}.

Data perturbation techniques preserve the original format of the input data or the output results. As a result of that, data perturbation techniques often allow some privacy leak ~\cite{xu2014building}. A privacy model defines and identifies the limits of private information protection and disclosure of a certain perturbation mechanism~\cite{machanavajjhala2015designing}. Earlier privacy models include $k-anonymity$, $l-diversity$, $(\alpha, k)-anonymity$, $t-closeness$~\cite{chamikara2019efficient2,li2007t}.  It was shown that these models are vulnerable to different attacks such as minimality attack ~\cite{zhang2007information}, composition based attacks~\cite{ganta2008composition}, and foreground knowledge~\cite{wong2011can}. These attacks exploit the perturbed data to reconstruct private information. Differential privacy (DP) is a strong privacy model that is trusted to provide a better level of privacy guarantee compared to previous privacy models ~\cite{dwork2009differential, mohammed2011differentially}.

There are different data perturbation approaches to achieve differential privacy. Laplace mechanism, Gaussian mechanism ~\cite{chanyaswad2018mvg}, geometric mechanism, randomized response ~\cite{qin2016heavy}, and staircase mechanisms ~\cite{kairouz2014extremal} are a few of the fundamental mechanisms used to achieve differential privacy. There are many practical examples where these fundamental mechanisms have been used to build differentially private algorithms. Differential Privacy for SQL Queries~\cite{johnson2018towards}, LDPMiner ~\cite{qin2016heavy}, PINQ~\cite{mcsherry2009privacy}, RAPPOR~\cite{erlingsson2014rappor}, Succinct histogram~\cite{bassily2015local} and Deep Learning with Differential Privacy~\cite{abadi2016deep} are a few examples of such practical applications.

 Literature shows a few attempts to address the issue of privacy leaks in deep learning algorithms by imposing private training ~\cite{abadi2016deep,shokri2015privacy,li2017multi,papernot2016semi,osia2017hybrid}.  Shokri, R. et al. ~\cite{shokri2015privacy} developed a distributed multi-party learning mechanism (referred to as  [SS15] in Table \ref{accomp})  for a neural network without sharing input datasets. They parallelized the learning process, which is based on the stochastic gradient descent optimization algorithm. In their method, the privacy loss is calculated per parameter of the model. This feature can entail a substantial privacy loss as there are many model parameters, often there can be thousands of such model parameters.  Abadi, M. et al. ~\cite{abadi2016deep} introduced an efficient differentially private mechanism (referred to as  [ACG+16] in Table \ref{accomp}) based on global differential privacy. Their model is capable of achieving high efficiency and performance under a modest privacy budget.  Their algorithm is based on a differentially private version of stochastic gradient descent, which runs on the TensorFlow software library for machine learning. Further, they introduced a mechanism to track privacy loss, the moments accountant, which allows tight automated analysis of privacy loss. But the additive bound $\delta$ of their ($\varepsilon$,$\delta$)-differential privacy mechanism may incur an unreliable level of privacy leak when the method is used for much larger datasets. Another shortcoming of the two methods [SS15] and [ACG+16] is the need for a trusted third party. Since both approaches are based on global differential privacy, the necessity of having a trusted third party cannot be avoided. This can be considered as a significant issue in applying these methods to real-world scenarios, where trusted curators are not always available. 

LATENT is designed to be aligned with machine-learning-as-a-service scenario, which has become popular due to the capabilities offered by large Internet-based companies, such as Google and Amazon ~\cite{abadi2016tensorflow}. For example Google's cloud-based machine learning engine provides the ability to build the models with multiple ML frameworks such as scikit-learn~\cite{pedregosa2011scikit}, XGBoost~\cite{chen2016xgboost}, Keras~\cite{chollet2015keras}, and TensorFlow~\cite{abadi2016tensorflow}. LATENT uses similar technologies for its implementation, replicating the technical settings of the environment offered by Google's cloud ML platform and other related services.

\section{Conclusion}
\label{conclusion}

We proposed a new local differentially private mechanism to train a deep neural network with high privacy and high accuracy.  Our model exhibits excellent accuracy even under extremely low privacy budgets (e.g. $\varepsilon=0.5$) compared to existing differentially private approaches. We achieve 95\%-96\% testing accuracy and 90\%-91\% testing accuracy for the MNIST dataset and CIFAR-10 dataset, respectively, with a high level of privacy (0.5-differential privacy). Due to the large feature space created by LATENT during the randomization process, it generates better accuracy for the CIFAR-10 dataset than the baseline CNN model without any privacy. Existing differentially private mechanisms are implemented using global differential privacy, and so they need a trusted curator. The untrusted curator setting of our approach provides a higher level of privacy while leaving a low level of computational burden to the data owners. Moving the convolutional module to the data owners produces additional privacy even without the application of randomization as the convolutional module output is a 1-D dimension-reduced output. The distribution of the CNN structure between data owners and servers also increases the flexibility of data processing in the big data context. This distribution also helps LATENT be easily adapted to innovations such as the amalgamation of SDN and NFV in edge-cloud interplay. When a large number of data owners communicate with a single server, the server has to be concerned only about generating the differentially private ANN model. The ability to use our method in the untrusted curator setting allows the private sharing of sensitive data and limits the privacy leak in distributed machine learning scenarios.  Since the proposed method is based on LDP, we do not make any architectural modifications to the fully connected artificial neural network component (which we call the FC module) of a convolutional network. Therefore, the input parameter selection (e.g. $\varepsilon$, $\alpha$, and the number of input bits) of the differentially private component (LATENT) is independent of the tuning processes (e.g. regularization, image augmentation, and hyperparameter tuning) of the FC module in the CNN architecture. This allows easy training and tuning of the FC module with a higher level of accuracy and an extreme level of privacy, resulting in an outstanding balance between privacy and utility.

Our approach opens up many future research directions. Investigating the possibility of reducing data sensitivity would be a good research avenue. Low sensitivity would allow the selection of an appropriate $\varepsilon$ value tailored to domain requirements.  We would also like to test our method on other deep learning architectures such as recurrent networks with LSTM (Long Short-Term Memory) and test it for other large datasets to find its performance and generalizability.

\appendices
\section{Proof of Unary Encoding (UE)}
\label{appdx1}

\begin{proof}
Considering a sensitivity of 2, choose $p$ and $q$ as follows,

\begin{equation}
   p = \frac{ e^{\frac{\varepsilon}{2}}}{1+ e^{\frac{\varepsilon}{2}}} 
   \label{prob1aa}
\end{equation}
\begin{equation}
   q = \frac{1}{1+ e^{\frac{\varepsilon}{2}}} 
   \label{prob2aa}
\end{equation}

\begin{equation}
\begin{aligned} \frac{\operatorname{Pr}\left[\boldsymbol{B} | v_{1}\right]}{\operatorname{Pr}\left[\boldsymbol{B} | v_{2}\right]} &=\frac{\prod_{i \in[d]} \operatorname{Pr}\left[\boldsymbol{B}[i] | v_{1}\right]}{\prod_{i \in[d]} \operatorname{Pr}\left[\boldsymbol{B}[i] | v_{2}\right]} \\ & \leq \frac{\operatorname{Pr}\left[\boldsymbol{B}\left[v_{1}\right]=1 | v_{1}\right] \operatorname{Pr}\left[\boldsymbol{B}\left[v_{2}\right]=0 | v_{1}\right]}{\operatorname{Pr}\left[\boldsymbol{B}\left[v_{1}\right]=1 | v_{2}\right] \operatorname{Pr}\left[\boldsymbol{B}\left[v_{2}\right]=0 | v_{2}\right]} \\ &=\frac{p}{q} \cdot \frac{1-q}{1-p}=e^{\varepsilon} \end{aligned}
\label{moueiproof}
\end{equation}
\end{proof}

\section{Proof of Optimized Unary Encoding (UE)}
\label{appdxoue}

\begin{proof}
Considering a sensitivity of 2, define,

\begin{equation}
   p = \frac{1}{2} 
   \label{prob1aa11}
\end{equation}
\begin{equation}
   q = \frac{1}{1+ e^{\varepsilon}} 
   \label{prob2aa22}
\end{equation}

\begin{equation}
\begin{aligned} \frac{\operatorname{Pr}\left[\boldsymbol{B} | v_{1}\right]}{\operatorname{Pr}\left[\boldsymbol{B} | v_{2}\right]} &=\frac{\prod_{i \in[d]} \operatorname{Pr}\left[\boldsymbol{B}[i] | v_{1}\right]}{\prod_{i \in[d]} \operatorname{Pr}\left[\boldsymbol{B}[i] | v_{2}\right]} \\ & \leq \frac{\operatorname{Pr}\left[\boldsymbol{B}\left[v_{1}\right]=1 | v_{1}\right] \operatorname{Pr}\left[\boldsymbol{B}\left[v_{2}\right]=0 | v_{1}\right]}{\operatorname{Pr}\left[\boldsymbol{B}\left[v_{1}\right]=1 | v_{2}\right] \operatorname{Pr}\left[\boldsymbol{B}\left[v_{2}\right]=0 | v_{2}\right]} \\ &=\frac{p}{q} \cdot \frac{1-q}{1-p}=e^{\varepsilon} \end{aligned}
\label{oueiproof2}
\end{equation}
\end{proof}

\section{Proof of the Upper Bound theorem}
\label{upbproof}
\begin{proof}
Considering a sensitivity of 2, choose $p$ and $q$ of Eq.\ref{pereq11} according to Eq.\ref{prob111} and Eq. \ref{prob222} respectively.
 
\begin{equation}
   p = \frac{\alpha e^{\frac{\varepsilon}{2}}}{1+\alpha e^{\frac{\varepsilon}{2}}} 
   \label{prob111}
\end{equation}
\begin{equation}
   q = \frac{1}{1+\alpha e^{\frac{\varepsilon}{2}}} 
   \label{prob222}
\end{equation}

According to Eq. \ref{pqratio}, we can write,

\begin{equation}
\begin{aligned}
\varepsilon=\ln \left(\frac{\left(\frac{\alpha e^{\frac{\varepsilon}{2}}}{1+\alpha e^{\frac{\varepsilon}{2}}} \right) \left(1-\frac{1}{1+\alpha e^{\frac{\varepsilon}{2}}}\right)}{\left(1-\frac{\alpha e^{\frac{\varepsilon}{2}}}{1+\alpha e^{\frac{\varepsilon}{2}}} \right) \left(\frac{1}{1+\alpha e^{\frac{\varepsilon}{2}}}\right)}\right) \\
\varepsilon = \ln (\alpha^2 e^{\varepsilon})
\end{aligned}
\label{deripqratio}
\end{equation}
Therefore, $UB(\varepsilon) = \ln (\alpha^2 e^{\varepsilon})$

\end{proof}

\section{Proof of modified OUE (MOUE)}
\label{proofmoue}
\begin{proof}
Considering a sensitivity of 2, let,
 
\begin{equation}
   p = \frac{1}{1+\alpha} 
   \label{prob11133}
\end{equation}
\begin{equation}
   q = \frac{1}{1+\alpha e^{\varepsilon}} 
   \label{prob22233}
\end{equation}

\begin{equation}
\begin{aligned} \frac{\operatorname{Pr}\left[\boldsymbol{B} | v_{1}\right]}{\operatorname{Pr}\left[\boldsymbol{B} | v_{2}\right]} &=\frac{\prod_{i \in[d]} \operatorname{Pr}\left[\boldsymbol{B}[i] | v_{1}\right]}{\prod_{i \in[d]} \operatorname{Pr}\left[\boldsymbol{B}[i] | v_{2}\right]} \\ & \leq \frac{\operatorname{Pr}\left[\boldsymbol{B}\left[v_{1}\right]=1 | v_{1}\right] \operatorname{Pr}\left[\boldsymbol{B}\left[v_{2}\right]=0 | v_{1}\right]}{\operatorname{Pr}\left[\boldsymbol{B}\left[v_{1}\right]=1 | v_{2}\right] \operatorname{Pr}\left[\boldsymbol{B}\left[v_{2}\right]=0 | v_{2}\right]} \\ &=\frac{\left(\frac{1}{1+\alpha}\right)}{\left(\frac{\alpha}{1+\alpha}\right)} \cdot \frac{\left(\frac{\alpha e^{\varepsilon}}{1+\alpha e^{\varepsilon}} \right)}{\left(\frac{1}{1+\alpha e^{\varepsilon}} \right)}=e^{\varepsilon} \end{aligned}
\label{moueproof}
\end{equation}
\end{proof}

\section{Proof of $\varepsilon$-LDP for MOUE for high sensitivities}
\label{mohsen}
\begin{proof}
Given that LATENT has a sensitivity of $r\times l$ (refer Section \ref{latentzsense}), the privacy budget ($\epsilon$) needs to be divided by the sensitivity for each bit as proven by RAPPOR. For MOUE, $\operatorname{Pr}\left[\boldsymbol{B}\left[v_{1}\right]=1 | v_{1}\right]=\frac{1}{1+\alpha}$ and $\operatorname{Pr}\left[\boldsymbol{B}\left[v_{1}\right]=1 | v_{2}\right]=\frac{\alpha}{1+\alpha}$. Hence, 

\begin{equation}
    \operatorname{Pr}\left[\boldsymbol{B}\left[v_{1}\right]=1 | v_{2}\right]=\frac{\alpha e^{\frac{\varepsilon}{rl}}}{1+\alpha e^{\frac{\varepsilon}{rl}}}
\end{equation}

\begin{equation}
    \operatorname{Pr}\left[\boldsymbol{B}\left[v_{2}\right]=0 | v_{2}\right]=\frac{1}{1+\alpha e^{\frac{\varepsilon}{rl}}}
\end{equation}

Therefore,

\begin{equation}
\begin{aligned} \frac{\operatorname{Pr}\left[\boldsymbol{B} | v_{1}\right]}{\operatorname{Pr}\left[\boldsymbol{B} | v_{2}\right]} &=\frac{\prod_{i \in[d]} \operatorname{Pr}\left[\boldsymbol{B}[i] | v_{1}\right]}{\prod_{i \in[d]} \operatorname{Pr}\left[\boldsymbol{B}[i] | v_{2}\right]} \\ & \leq \left( \frac{\operatorname{Pr}\left[\boldsymbol{B}\left[v_{1}\right]=1 | v_{1}\right] \operatorname{Pr}\left[\boldsymbol{B}\left[v_{2}\right]=0 | v_{1}\right]}{\operatorname{Pr}\left[\boldsymbol{B}\left[v_{1}\right]=1 | v_{2}\right] \operatorname{Pr}\left[\boldsymbol{B}\left[v_{2}\right]=0 | v_{2}\right]} \right)^{rl} \\ &=\left(\frac{\left(\frac{1}{1+\alpha}\right)}{\left(\frac{\alpha}{1+\alpha}\right)} \cdot \frac{\left(\frac{\alpha e^{\frac{\varepsilon}{rl}}}{1+\alpha e^{\frac{\varepsilon}{rl}}} \right)}{\left(\frac{1}{1+\alpha e^{\frac{\varepsilon}{rl}}} \right)}\right)^{rl}=e^{\varepsilon} \end{aligned}
\label{repproof}
\end{equation}

\end{proof}

\section{Proof of the $\varepsilon$-LDP of the utility enhancing randomization step of LATENT}

\label{proofuer}
\begin{proof}
Considering a sensitivity of $rl$, choose the randomization probabilities according to Eq. \ref{twomodprobqq}.
\begin{equation}
p(B[i]v)=\left\{\begin{array}{ll}{\operatorname{Pr}\left[\boldsymbol{B}\left[v_{1}\right]=1 | v_{1}\right] = \frac{\alpha}{1+\alpha}} & {\text { if } i \in 2n; n\in \field{N}} \\
{\operatorname{Pr}\left[\boldsymbol{B}\left[v_{2}\right]=0 | v_{1}\right] = \frac{\alpha e^{\frac{\varepsilon}{rl}}}{1+\alpha e^{\frac{\varepsilon}{rl}}}} & {\text ~~~~~~~ \ditto} \\ {\operatorname{Pr}\left[\boldsymbol{B}\left[v_{1}\right]=1 | v_{1}\right] = \frac{1}{1+\alpha^3}} & {\text { if } i \in 2n+1} \\ {\operatorname{Pr}\left[\boldsymbol{B}\left[v_{2}\right]=0 | v_{1}\right] = \frac{\alpha e^{\frac{\varepsilon}{rl}}}{1+\alpha e^{\frac{\varepsilon}{rl}}}} & {\text ~~~~~~~ \ditto }\end{array}\right.
\label{twomodprobqq}
\end{equation}

\begin{equation}
\begin{aligned} \frac{\operatorname{Pr}\left[\boldsymbol{B} | v_{1}\right]}{\operatorname{Pr}\left[\boldsymbol{B} | v_{2}\right]} &= \frac{\prod_{i \in[d]} \operatorname{Pr}\left[\boldsymbol{B}[i] | v_{1}\right]}{\prod_{i \in[d]} \operatorname{Pr}\left[\boldsymbol{B}[i] | v_{2}\right]} \\
&=\frac{\prod_{i \in 2n} \operatorname{Pr}\left[\boldsymbol{B}[i] | v_{1}\right]}{\prod_{i \in 2n} \operatorname{Pr}\left[\boldsymbol{B}[i] | v_{2}\right]}\times \frac{\prod_{i \in 2n+1} \operatorname{Pr}\left[\boldsymbol{B}[i] | v_{1}\right]}{\prod_{i \in 2n+1} \operatorname{Pr}\left[\boldsymbol{B}[i] | v_{2}\right]} \\ 
& \leq \left( \frac{\operatorname{Pr}\left[\boldsymbol{B}\left[v_{1}\right]=1 | v_{1}\right] \operatorname{Pr}\left[\boldsymbol{B}\left[v_{2}\right]=0 | v_{1}\right]}{\operatorname{Pr}\left[\boldsymbol{B}\left[v_{1}\right]=1 | v_{2}\right] \operatorname{Pr}\left[\boldsymbol{B}\left[v_{2}\right]=0 | v_{2}\right]} \right)^{\frac{rl}{2}}_{i \in 2n}\times \\
&\left( \frac{\operatorname{Pr}\left[\boldsymbol{B}\left[v_{1}\right]=1 | v_{1}\right] \operatorname{Pr}\left[\boldsymbol{B}\left[v_{2}\right]=0 | v_{1}\right]}{\operatorname{Pr}\left[\boldsymbol{B}\left[v_{1}\right]=1 | v_{2}\right] \operatorname{Pr}\left[\boldsymbol{B}\left[v_{2}\right]=0 | v_{2}\right]} \right)^{\frac{rl}{2}}_{i \in 2n+1} \\
&=\left(\frac{\left(\frac{\alpha}{1+\alpha}\right)}{\left(\frac{1}{1+\alpha}\right)} \cdot \frac{\left(\frac{\alpha e^{\frac{\varepsilon}{rl}}}{1+\alpha e^{\frac{\varepsilon}{rl}}} \right)}{\left(\frac{1}{1+\alpha e^{\frac{\varepsilon}{rl}}} \right)}\right)^{\frac{rl}{2}}\left(\frac{\left(\frac{1}{1+\alpha^3}\right)}{\left(\frac{\alpha^3}{1+\alpha^3}\right)} \cdot \frac{\left(\frac{\alpha e^{\frac{\varepsilon}{rl}}}{1+\alpha e^{\frac{\varepsilon}{rl}}} \right)}{\left(\frac{1}{1+\alpha e^{\frac{\varepsilon}{rl}}} \right)}\right)^{\frac{rl}{2}}\\
&=e^{\varepsilon} \end{aligned}
\end{equation}

\end{proof}

\ifCLASSOPTIONcaptionsoff
  \newpage
\fi

\bibliographystyle{IEEEtran}



\end{document}